\documentclass{article}

\usepackage[preprint]{neurips_2026}

\usepackage[utf8]{inputenc} % allow utf-8 input
\usepackage[T1]{fontenc}    % use 8-bit T1 fonts
\usepackage{hyperref}       % hyperlinks
\usepackage{url}            % simple URL typesetting
\usepackage{booktabs}       % professional-quality tables
\usepackage{amsfonts}       % blackboard math symbols
\usepackage{nicefrac}       % compact symbols for 1/2, etc.
\usepackage{microtype}      % microtypography
\usepackage{xcolor}         % colors

\usepackage{algorithm}
\usepackage{algpseudocode}
\usepackage{amsmath}
\usepackage{graphicx}

\usepackage{multirow}

\usepackage[most]{tcolorbox}
\usepackage{listings}
\usepackage{xcolor}
\usepackage{fvextra}

\usepackage{siunitx}
\usepackage{fancyvrb}
\usepackage{placeins}

\usepackage{longtable}
\usepackage{multirow}
\usepackage{booktabs}
\usepackage{amssymb} % for \checkmark
\usepackage{tabularx}
 
\definecolor{revisionblue}{RGB}{70, 120, 180} %for tracking changes

\title{Discovering Structurally Plausible and Interpretable Cognitive Models with Large Language Models}

\author{%
Boyoung Kwon$^{1}$,
Jiyoon Seo$^{2}$,
Sang Ho Lee$^{1,2,3\,\dagger}$
\\
$^{1}$Graduate School of Data Science, KAIST
\\
$^{2}$School of Digital Humanities and Computational Social Sciences, KAIST
\\
$^{3}$Department of Brain \& Cognitive Sciences, KAIST
\\
\texttt{\{bykwon, cristina0106, sangholee\}@kaist.ac.kr}
}

\begin{document}

\maketitle
\begingroup
\renewcommand{\thefootnote}{\ensuremath{\dagger}}
\footnotetext{Corresponding author.}
\endgroup

\begin{abstract}
  % The abstract paragraph should be indented \nicefrac{1}{2}~inch (3~picas) on both the left- and right-hand margins. Use 10~point type, with a vertical spacing (leading) of 11~points. The word \textbf{Abstract} must be centered, bold, and in point size 12. Two line spaces precede the abstract. The abstract must be limited to one paragraph.

  Computational cognitive modeling has traditionally relied on manual model construction, which is a time-intensive process that requires substantial domain expertise. Although recent LLM-based approaches have begun to automate model generation, they often prioritize predictive accuracy at the expense of structural plausibility and parameter identifiability. We introduce DisCo, an LLM-guided framework that discovers interpretable symbolic cognitive models through evolutionary search with two structural components: a Validator that filters structurally invalid models, and a Cleaner that promotes parameter identifiability by removing redundant parameters. Across three behavioral tasks, DisCo consistently discovers models that achieve performance competitive with task-specific baselines. Ablation analyses show that the Validator is important for recovering psychologically meaningful structure. Without this component, models tend to violate behavioral constraints and show less consistent associations with external cognitive measures. These results suggest that LLM-driven model discovery, when coupled with appropriate structural constraints, can yield cognitively plausible models that go beyond predictive fit.

\end{abstract}

\section{Introduction}
Computational cognitive models have long served as a central tool for explaining human behavior in terms of latent cognitive mechanisms. By formalizing psychological processes as mathematical equations, these models allow researchers to quantify individual differences in cognition and predict behavior under novel conditions. However, developing such models requires substantial domain expertise, coding proficiency, and extensive trial-and-error refinement \citep{agrawal2020scaling, musslick2025automating}. This manual process is not only time-consuming but also vulnerable to researchers' theoretical preconceptions, which may constrain exploration to a narrow subset of possible model structures.

These limitations have motivated a growing interest in automated approaches to cognitive model discovery. In particular, large language models (LLMs) have emerged as powerful tools for automating scientific discovery, with demonstrated applications ranging from equation discovery in physics to symbolic regression in biology \citep{Romera-Paredes2024, shojaee2025llmsr}. Building on these advances, researchers have begun applying LLMs specifically to cognitive model generation. Recent frameworks such as GeCCo \citep{rmus2025generating} and CogFunSearch \citep{castro2025discovering} have demonstrated that LLMs can generate cognitive models with strong predictive performance, primarily evaluated by their fit to behavioral data.

However, predictive accuracy alone is insufficient for the scientific goals of cognitive modeling. A model may fit behavioral data well while violating the causal structure of the task, for example by using future outcomes that would not be available at the time of decision. Such a model may achieve strong quantitative performance while misrepresenting the underlying cognitive processes it is intended to capture. Moreover, if model parameters are to be interpreted as markers of cognitive traits, such as learning rate or response sensitivity, they should be identifiable from behavioral data \citep{wilson2019ten}. Goodness of fit is an unreliable indicator of parameter identifiability, as multiple parameter configurations can produce indistinguishable outputs.

% Despite these advances, predictive accuracy alone is insufficient for the scientific goals of cognitive modeling. In cognitive modeling, parameters are interpreted as quantitative markers of cognitive traits (e.g., impulsivity, learning rate) \citep{wilson2019ten}, and should therefore be uniquely identifiable and correspond to meaningful cognitive constructs. Moreover, models that achieve high predictive fit through scientifically invalid means may go undetected, as such violations are not apparent from predictive accuracy alone. For instance, a model that references future reward structures unavailable at decision time can yield strong fits while fundamentally misrepresenting the underlying cognitive process.

We introduce \textbf{DisCo} (Discovery of Interpretable Symbolic Cognitive Models), an LLM-guided framework that incorporates two structural components into the discovery loop. First, a Validator filters models that violate causal assumptions of the task, such as referencing future outcomes unavailable at decision time. Second, a Cleaner promotes parameter identifiability by detecting and removing redundant parameters based on posterior covariance structure. Together, these components promote models that are not only predictive but also structurally plausible and interpretable.

We evaluate DisCo across three well-established behavioral tasks representing distinct cognitive functions spanning prospective, adaptive, and retrospective processes: planning (two-step task; \citealt{daw2011model}), learning (probabilistic reversal learning task; \citealt{zorowitz2023inattentive}), and memory (recognition memory task; \citealt{cataldo2023abnormal}). Our results show that explicit validation is critical for recovering psychologically meaningful model structure.

\section{Related Work}

\subsection{Automated Scientific Discovery}
Automated scientific discovery has traditionally relied on symbolic regression methods, particularly genetic programming (GP) \citep{koza1992genetic, schmidt2009distilling}. GP evolves candidate mathematical expressions through evolutionary operations, but faces challenges in scalability and incorporating prior knowledge \citep{petersen2019deep, makke2024interpretable}.

More recently, the emergence of large language models (LLMs) has opened new avenues for automated scientific discovery. FunSearch \citep{Romera-Paredes2024} combines LLMs with evolutionary search to support mathematical discoveries, producing interpretable programs rather than black-box solutions. LLM-SR \citep{shojaee2025llmsr} demonstrates that LLMs' scientific priors can guide equation discovery with superior generalization, while EUREKA \citep{ma2023eureka} achieves human-level performance in reward function design through LLM-based evolutionary optimization. These developments position LLMs as powerful tools for automated, interpretable scientific discovery.

\subsection{Deep Learning in Cognitive Modeling}
In cognitive modeling, deep learning approaches have achieved strong predictive performance in complex behavioral tasks \citep{dezfouli2019models}. Deep neural networks can serve as a flexible framework for developing cognitive models, but their black-box nature limits their utility for theory-driven cognitive modeling. Several methods address this limitation, including disentangled representations via information bottlenecks \citep{miller2023rnns}, simplified architectures with minimal units \citep{ji2025discovering}, and hybrid neural–cognitive models \citep{eckstein2023predictive}. However, such approaches often rely on predefined architectural constraints or model structures, limiting their ability to automatically discover novel cognitive mechanisms.

\subsection{Automated Discovery of Cognitive Models}
Recent frameworks leverage LLMs to generate cognitive models that explain behavioral data. CogFunSearch \citep{castro2025discovering} adapts the FunSearch \citep{Romera-Paredes2024} framework by utilizing an LLM as a mutation operator to evolve symbolic programs that maximize the likelihood of observed behavior. Similarly, GeCCo \citep{rmus2025generating} employs in-context learning to translate task descriptions directly into computational models. These approaches primarily evaluate discovered models via goodness-of-fit and qualitative theoretical alignment, with less attention paid to the plausibility of the model structure or to quantitative indices of theoretical alignment. We advance this paradigm by incorporating additional evaluation steps typically performed by human cognitive modelers to improve the interpretability of the model: assessing model structure for parameter identifiability and causal consistency, and correlating parameter estimates with external indices of cognitive traits.

\section{Method}

\subsection{Dataset}

\begin{figure}
    \centering
    \includegraphics[width=\columnwidth]{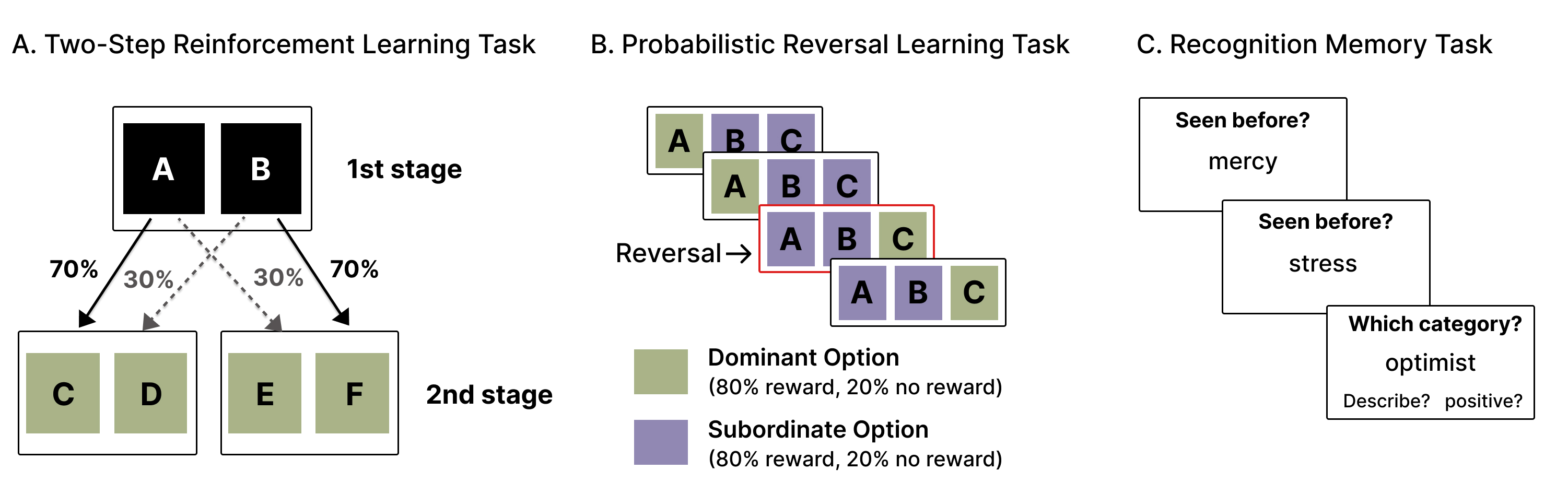}
    \caption{Overview of the behavioral tasks. We utilize large-scale data from (A) the two-step reinforcement learning task, (B) the probabilistic reversal learning task, and (C) the recognition memory task.}
    \label{fig:dataset}
\end{figure}

To evaluate our framework, we employed three behavioral datasets that meet the following criteria: (i) sufficiently large sample sizes, and (ii) availability of well-established baseline computational models. Each dataset primarily engages a distinct cognitive domain, namely planning, learning, and memory, allowing us to assess whether the discovered model generalizes across qualitatively different cognitive functions. Although no single task isolates one process entirely, each paradigm places relatively greater demand on one domain. An overview of the task paradigms is presented in Figure~\ref{fig:dataset}. Additional details on each dataset, including the baseline models and external measures, are provided in Appendix~\ref{app:details}.

\subsubsection{Planning}
\label{sec:dataset_twostep}
 
 \paragraph{Task.} The two-step reinforcement learning task is a widely used paradigm for studying planning. We used the dataset from \citet{gillan2016characterizing}, which is based on the paradigm originally introduced by \citet{daw2011model}. The task was designed to distinguish between habitual (model-free) and goal-directed (model-based) decision-making processes. On each trial, participants chose between two abstract stimuli at the first stage, which led probabilistically to one of two second-stage states. At the second stage, participants selected between two additional stimuli, after which a monetary outcome was delivered probabilistically. We used the trial-by-trial data from Experiment 1 ($N = 548$ participants, 200 trials each).

\paragraph{Baseline Model.} 
% 줄인 버전
For the two-step reinforcement learning task, we used simplified versions of two models from \citet{daw2011model} as baselines: a model-free 
temporal-difference (TD) learning model that updates action values solely from experienced rewards, and a hybrid model that additionally incorporates model-based planning via the task's transition structure.

% For the two-step reinforcement learning task, we considered simplified versions of two baseline models from \citet{daw2011model}. The first is a model-free temporal-difference (TD) learning model that maintains separate action values for the first and second stages, updating first-stage values via credit assignment from second-stage outcomes without representing the task's transition structure.
% The second is a hybrid model that integrates model-free learning with model-based planning; first-stage values are computed as a weighted combination of cached model-free values and model-based values obtained via one-step look-ahead using the known transition probabilities.
% Both models use softmax action selection and TD learning for value updates.
% Full specifications are provided in Appendix~\ref{app:twostep}.

% Revised (0505) 
\subsubsection{Learning}

\paragraph{Task.}
We used the dataset from \citet{zorowitz2023inattentive}, which is based on a probabilistic reversal learning task designed to probe reinforcement-learning processes. On each trial, participants chose among three options and received probabilistic feedback. The dominant option yielded a reward with 80\% probability, while the subordinate options each yielded 20\%. The dominant option changed every 15 trials, requiring flexible adaptation. We used the trial-by-trial data from $N = 386$ participants (90 trials each).

\paragraph{Baseline Model.}
A temporal-difference (TD) learning model \citep{mihatsch2002risk} served as the baseline. The model updates expected values for each option using asymmetric learning rates for positive and negative prediction errors, with action selection governed by a softmax rule. 

\subsubsection{Memory}

\paragraph{Task.}
The dataset from \citet{cataldo2023abnormal} is based on an emotional memory task designed to investigate valence-dependent memory biases. During the encoding phase, participants judged normatively positive and negative words in either a self-reference (``Does this word describe you?'') or valence (``Is this word positive?'') task. During the test phase, participants viewed an equal mix of studied (old) and unstudied (new) words and judged whether each had been seen before. We used trial-by-trial data from $N = 1{,}358$ participants (approximately 100 recognition trials each).

\paragraph{Baseline Model.}
We used a trial-level equal-variance signal detection theory (SDT) model \citep{macmillan2005detection} as the baseline.  The model treats each trial's binary recognition response as determined by the observer's ability to discriminate old from new items and their response criterion. 

% The model treats each trial's binary response as a Bernoulli random variable determined by discriminability ($d'$) and criterion ($c$), resulting in two free parameters.

\subsection{Problem Formulation}
\label{sec:problem}

We frame cognitive model discovery as the joint optimization of model structure and parameters.

\paragraph{Behavioral Data.} Consider a dataset $\mathcal{D} = \{d_1, \dots, d_N\}$ consisting of behavioral trajectories from $N$ subjects. For each subject $i$, the data $d_i = \{(s_{i,t}, c_{i,t}, r_{i,t})\}_{t=1}^{T_i}$ consists of a sequence of (possibly implicit) states $s_{i,t} \in \mathcal{S}$, choices $c_{i,t} \in \mathcal{A}$, and rewards $r_{i,t} \in \mathbb{R}$ over $T_i$ trials. 

\paragraph{Symbolic Cognitive Model.} We seek to discover a symbolic cognitive model $\phi$ from a space of candidate models $\Phi$. Crucially, each model $\phi$ defines not only the executable function $f_\phi$ but also the valid parameter space $\Theta_\phi \subset \mathbb{R}^k$ (e.g., bound constraints generated by the LLM). Given subject-specific parameters $\theta_i \in \Theta_\phi$, the model maps the interaction history $h_{i,t}$ to a probability distribution over the next action:
\begin{equation}
    P(c_{i,t} \mid h_{i,t}; \phi, \theta_i) = f_\phi(h_{i,t}; \theta_i)
\end{equation}

\paragraph{Optimization Objective.} Since the datasets considered in this work involve sequential learning tasks, we employ a chronological split validation strategy to preserve temporal order and to capture trial-by-trial learning dynamics. 
For each participant $i$, we partition the data $d_i$ into the first half $d_i^{\mathrm{train}}$ for parameter estimation and the subsequent half $d_i^{\mathrm{val}}$ for model evaluation. 
We use the Bayesian Information Criterion (BIC) as the model-selection objective. BIC balances goodness of fit against model complexity by imposing a penalty for the number of free parameters; lower BIC scores indicate a better trade-off between predictive fit and parsimony \citep{schwarz1978estimating}.
We seek the program $\phi^*$ that minimizes the average BIC evaluated on the validation segment:
\begin{equation}
    \label{eq:objective}
    \phi^* = \arg\min_{\phi \in \Phi} \frac{1}{N} \sum_{i=1}^{N} \mathrm{BIC}(\phi, \hat{\theta}_i; d_i^{\mathrm{val}})
\end{equation}
where $\hat{\theta}_i$ are the parameters estimated from the training segment $d_i^{\mathrm{train}}$ via constrained maximum likelihood estimation (MLE) subject to $\theta_i \in \Theta_\phi$. Note that while parameters are fixed after the training phase, the model simulation proceeds continuously to preserve latent state dynamics (e.g., Q-values) accumulated during training when evaluating on $d_i^{\mathrm{val}}$. 
This within-participant chronological split was used during model discovery on the training participants.
% final performance is reported on held-out test participants (see Appendix~\ref{app:dataset}).
For final evaluation, the selected program $\phi^*$ was applied to held-out test participants using the same procedure (see Appendix~\ref{app:dataset} for details of the data-splitting procedure and alternative strategies supported by DisCo).%: model parameters were independently estimated from the first half of each test participant's data, and BIC was computed on the second half, with latent state dynamics carried forward across the split.% (see Appendix~\ref{app:dataset}).

\paragraph{Parameter Identifiability and Significance.} 
For a discovered model to be interpretable, its parameters must be identifiable and influential. We define two criteria for valid parameters: 
(i) Multicollinearity: Parameters must not exhibit excessive multicollinearity (e.g., $|\rho| < \tau_{\mathrm{corr}}$), which ensures that each parameter represents a distinct cognitive trait. 
(ii) Individual Variability: Parameters must demonstrate meaningful variance across subjects ($\sigma^2 > \tau_{\mathrm{var}}$), preventing the inclusion of constants that fail to capture behavioral heterogeneity. 
Accordingly, we adopt these criteria for evaluating discovered models, such that only scientifically plausible models are considered valid candidates.

\paragraph{Association with Cognitive Traits.}
In addition to predictive accuracy and parameter validity, we consider whether fitted parameters exhibit external cognitive relevance. When independent psychological measures $\mathcal{M}$ are available, we compute correlations between fitted parameter estimates and each measure in $\mathcal{M}$. To account for multiple comparisons across parameter--measure pairs, we apply the Benjamini--Hochberg false discovery rate (FDR) correction. 
% We regard an association as externally meaningful when $|\mathrm{Corr}(\hat{\theta}, \mathcal{M})| > \delta$ and $p_{\mathrm{FDR}} < 0.05$, where $\delta$ denotes a predefined minimum effect-size threshold. This criterion provides complementary evidence of cognitive relevance beyond predictive fit.
We regard an association as externally meaningful when it survives Benjamini--Hochberg FDR correction at $p_{\mathrm{FDR}} < 0.05$. We report the corresponding correlation coefficients as effect-size estimates.

\begin{figure}
    \centering
    \includegraphics[width=\linewidth]{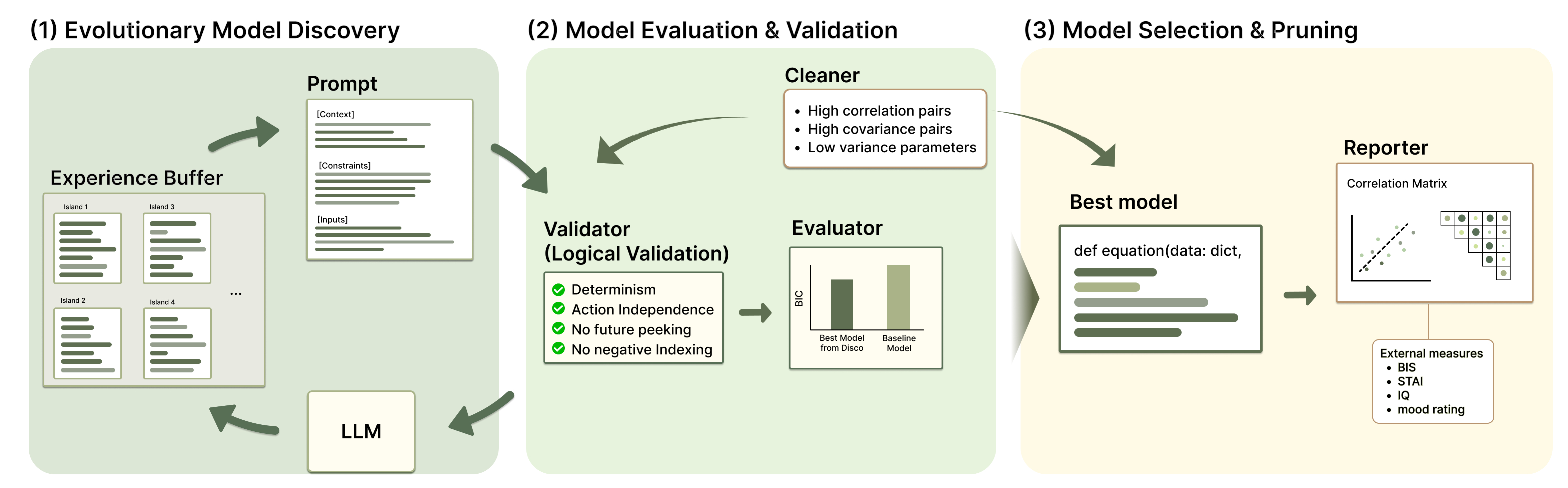}
    \caption{Overview of the DisCo pipeline.}
    \label{fig:pipeline}
\end{figure}

\subsection{The DisCo Framework}
Building upon the framework of LLM-based evolutionary algorithms \citep{shojaee2025llmsr, Romera-Paredes2024}, we propose DisCo (Discovery of Interpretable Symbolic Cognitive Models). As illustrated in Figure~\ref{fig:pipeline}, this approach leverages the code-generation capabilities of LLMs within an evolutionary strategy to explore the space of possible cognitive equations. DisCo specifically tailors this methodology to automate the scientific discovery process, seeking a model $\phi^*$ that optimizes predictive accuracy and scientific interpretability.

The optimization process is formulated as an island-based evolutionary algorithm where the LLM functions as a mutation and crossover operator. The pipeline operates through iterative evolutionary cycles and a final refinement stage: (1) \textit{Evolutionary Hypothesis Generation} generates candidate programs; (2) \textit{Model Validation} filters invalid models using logical integrity checks and chronological validation; and (3) \textit{Model Selection \& Pruning} periodically removes redundant parameters to encourage parsimony (during island updates) and, where applicable, to assess the resulting model based on external cognitive measures. The full discovery procedure is summarized in Appendix~\ref{app:algorithm}. 

\subsubsection{Evolutionary Model Discovery}
\label{sec:discovery}

\paragraph{Sampler (LLM-based Mutation).}
To bootstrap the discovery process, we initialize all island buffers with an initial exemplar provided in the template. Starting from these seeds, we leverage the code-generation capabilities of LLMs to evolve model structures. The Sampler constructs a prompt (see Appendix~\ref{app:prompt}) containing the task specification (function signature and state variables) and a set of high-scoring exemplars sampled from the current island's experience buffer. The LLM outputs executable Python code defining the forward pass and parameter bounds.  

\paragraph{Experience Buffer.}
We maintain diversity using an island model approach \citep{Romera-Paredes2024}. Each island buffer $B_k$ stores a population of programs sorted by their BIC scores. Periodically, the pipeline performs a migration step where top-performing models (i.e., those with the lowest BIC) are exchanged between islands to prevent premature convergence to local optima.

\subsubsection{Model Evaluation and Validation}
\label{sec:eval}

\paragraph{Validator (Logical Validation).}

Before fitting, generated programs undergo strict causal validation to filter out scientifically invalid models. We encourage four integrity constraints to help ensure that generated models respect the causal structure of sequential decision-making:

(i) Determinism: Given identical inputs, the model must produce identical outputs. Formally, for any history $h_{i,t}$ and parameters $\theta_i$, repeated evaluations must satisfy $P(c_{i,t} \mid h_{i,t}; \phi, \theta_i) = P'(c_{i,t} \mid h_{i,t}; \phi, \theta_i)$. This prevents the use of stochastic operations (e.g., \texttt{np.random}) within the model.

(ii) Independence of Action: The probability $P(c_{i,t} \mid h_{i,t}; \phi, \theta_i)$ must be calculated without accessing the actual choice $c_{i,t}$. This helps ensure the model predicts choice probabilities before observing the action, preventing trivial solutions that simply copy the observed choice.

(iii) No Future Peeking: Predictions at time $t$ must not change when future data ($t+k$, where $k > 0$) is perturbed. We verify this through randomized causal fuzzing: for randomly selected timesteps $t$, we corrupt future observations and confirm that $P(c_{i,t} \mid h_{i,t})$ remains invariant.

(iv) No Negative Indexing: Array access with negative indices (e.g., \texttt{data[-1]}) is disallowed to prevent cyclic information leakage, where the first prediction could depend on the final observation.

While these constraints specifically address causal validity in sequential tasks, the proposed validation mechanism is customizable, allowing users to specify alternative integrity checks appropriate for their specific domain directly within the code template.

\paragraph{Evaluator (Optimization \& Scoring).}
The Evaluator module quantifies the quality of candidate models that pass the logical validation. It solves the optimization problem defined in Equation~\ref{eq:objective}. Consistent with the chronological split strategy, we estimate the parameters $\theta$ on $d_i^{\text{train}}$ using the L-BFGS-B algorithm, with parameter estimates being bounded by LLM-generated constraints.
The goodness-of-fit score is then computed as the average of BIC scores on the validation set $d_i^{\text{val}}$. %Further details on the evaluation procedure are provided in Appendix~\ref{app:dataset}.

\subsubsection{Model Selection and Pruning}
\label{sec:selection}

\paragraph{Cleaner (Parameter Pruning).}
To promote parsimony, the Cleaner module periodically inspects the population for over-parameterization. We identify problematic parameters based on two criteria: (i) Multicollinearity ($|\rho| < \tau_{\text{corr}}$) and (ii) Individual variability ($\sigma^2 < \tau_{\text{var}}$). When detected, the LLM is prompted (Appendix~\ref{app:prompt_cleaner}) to simplify the equation by removing or consolidating the redundant parameters. If the simplified model $\phi'$ maintains a comparable BIC score to the original $\phi$, it replaces the complex version in the buffer.

% 수정한 부분
\paragraph{Reporter (External Validation).}
The Reporter module applies the external association criterion defined above to the top candidate models from the evolutionary search. Specifically, it correlates the fitted parameter estimates $\hat{\theta}$ of each candidate model with independent external measures $\mathcal{M}$ and summarizes which parameter--measure associations satisfy the FDR-corrected significance criterion. These results are used to characterize and compare candidate models in terms of external cognitive relevance, complementing the primary selection based on validation BIC.

\section{Experiments} 

\subsection{Experimental Setup}
We evaluated the performance of DisCo-generated models against established baseline computational models across three tasks: planning, learning, and memory. For the model discovery process, we used Qwen2.5-32B \citep{qwen2.5}. All LLM inference was performed using a vLLM server \citep{kwon2023efficient} running on two NVIDIA RTX A6000 GPUs (48GB VRAM each). 

% To balance exploration and precision, we employed a temperature setting of $T=0.6$ during the evolutionary generation phase, encouraging diversity in candidate programs. 
We used $T=0.6$ during evolutionary generation to encourage candidate diversity and $T=0.0$ during the Cleaner phase for deterministic pruning. 
Participants in each task were split into training and held-out test sets at an 8:2 ratio. % (see Appendix~\ref{app:dataset} for details).
For each run, DisCo generated and evaluated 200 candidate models, selected the final model on the training set, and computed BIC on the held-out test set. We repeated all experiments with five random seeds to assess the stability of the discovery process. Additional configuration information is provided in Appendix~\ref{app:config}.

\subsection{Performance Across Cognitive Domains}

\begin{table}
\centering
\small
\caption{Comparison of BIC scores across tasks, evaluated on held-out test participants. For DisCo, results are reported across five independent runs (mean $\pm$ std and range). For the baseline models, BIC was computed once and is reported as a single score.}
\label{tab:main_results}
\begin{tabular}{llccc}
\toprule
\textbf{Task} & \textbf{Model} & \textbf{BIC Mean $\pm$ Std} & \textbf{BIC range} \\
\midrule
\multirow{5}{*}{Planning} 
  & Model-Free       & 257.89        & --   \\
  & Hybrid           & 255.22        & --   \\
  & DisCo (Null)     & $240.54 \pm 8.07$  & [229.38, 249.36]   \\
  & DisCo (MF)       & $221.16 \pm 1.75$  & [218.27, 223.57]   \\
  & DisCo (Hybrid)   & $221.32 \pm 2.83$  & [216.20, 224.94]   \\
\midrule
\multirow{3}{*}{Learning} 
  & TD Learning      & 76.06        & --   \\
  & DisCo (Null)     & $70.64 \pm 0.15$ &  [70.57, 70.94] \\
  & DisCo (TD Learning)& $74.49 \pm 2.20$ & [70.27, 76.71]   \\
  
\midrule
\multirow{3}{*}{Memory} 
  & SDT              & 61.81        & --   \\
  & DisCo (Null)     & $51.95 \pm 0.04$ & [51.93, 52.02]   \\
  & DisCo (SDT)      & $56.03 \pm 1.93$ & [53.46, 59.28]   \\
\bottomrule
\end{tabular}
\end{table}

To evaluate DisCo, we applied the full pipeline to the three behavioral datasets. For each dataset, we compared BIC scores between DisCo-generated models and task-specific baselines: a standard model-free/hybrid RL model for the planning task, TD learning for the learning task, and SDT for the memory task. In addition to seeding the search with these task-specific baseline models, we also included a null-seeded condition, where DisCo was initialized from a minimal model with no task-specific computational assumptions. All DisCo runs used identical hyperparameters across tasks, with prompts reported in Appendix~\ref{app:prompt}.

%Table~\ref{tab:main_results} summarizes BIC scores across five independent runs with different random seeds. 
% Revised (260505)
% Table~\ref{tab:main_results} summarizes BIC scores across five random seeds. 
% DisCo consistently discovers models that outperform the baselines on the planning task and the learning task, with low variance across runs indicating stable convergence. %On the memory task, DisCo achieves comparable performance to the SDT baseline. 
% On the memory and learning tasks, DisCo-discovered models achieve competitive or better test-set BIC scores compared with the corresponding task-specific baselines. 
% Notably, the null-seeded DisCo condition also performs strongly on these tasks, suggesting that the LLM-guided search process can identify compact task-relevant model structures even without being initialized from a strong domain-specific baseline. 
Table~\ref{tab:main_results} summarizes BIC scores across five random seeds. Lower scores indicate better predictive fit. For DisCo variants, the labels in parentheses indicate the baseline model used to initialize the search; for example, DisCo (MF) denotes initialization from the model-free baseline, whereas DisCo (Null) denotes initialization from a minimal null model without task-specific assumptions. DisCo consistently discovers models that match or outperform task-specific baselines across all three tasks, with low variance across runs indicating stable convergence. DisCo (Null) also achieves competitive performance on the memory and learning tasks, demonstrating that the LLM-guided search can identify compact model structures without domain-specific initialization. However, baseline seeding provided a clear advantage on the planning task, where DisCo (MF) and DisCo (Hybrid) substantially outperformed DisCo (Null), suggesting that established cognitive mechanisms can serve as useful structural priors when available. 

Beyond predictive performance, the discovered models may provide alternative accounts of the cognitive processes represented by the baseline models or extend those accounts with additional mechanisms. The learning task provides a clear example. The TD baseline model explains individual differences in reversal learning through separate learning rates for positive and negative prediction errors. In contrast, the discovered model uses a single learning rate and decay of unchosen-option values. It attributes individual differences to forgetting of unchosen options rather than differential sensitivity to positive and negative outcomes. This account predicts that, after a reversal, participants should be slower to choose the newly advantageous option the longer it has remained unchosen. The best discovered models for all three tasks and their differences from the corresponding baselines are described in detail in Appendix~\ref{app:best_model}.
%learning 예시를 본문에 추가. -> 이건 예시고 full description이 appendix에 있다고 언급하기

%The TD learning model's separate learning rates for positive and negative prediction errors are replaced by a single learning rate and decay of unchosen-option values. This suggests an alternative account of individual differences in observed behavior, attributing them to variation in a single learning rate and in forgetting of information, rather than to separate sensitivities to positive and negative prediction errors.

%% Preprint: rebuttal 이후 변경 내용
\subsection{Comparison with Existing LLM-based Model Discovery}
\label{sec:gecco_comparison}

To compare DisCo with existing LLM-based approaches to model discovery, we conducted a direct comparison with GeCCo \citep{rmus2025generating} across all three tasks. For each task, both methods were initialized with the same task-specific baseline model and evaluated under the same computational budget. Five independent runs were performed for each method, with 200 candidate models generated per run. The final model selected from each run was evaluated on held-out data. %, as described in Appendix~\ref{app:dataset}.
For this comparison, we define a final model as valid if it satisfies our post-hoc validity checks, which include checks of the structural constraints performed by the Validator as well as basic mathematical and code-level checks required for valid likelihood computation. % 문장 추가: invalid / valid 정의 추가
Table~\ref{tab:gecco_comparison} summarizes the held-out BIC scores across valid runs.

\begin{table}[t]
\centering
\small
\caption{Held-out BIC comparison between DisCo and GeCCo across tasks.
Statistics are computed over runs in which the final selected model
satisfied the validation criteria.}
\label{tab:gecco_comparison}
\begin{tabular}{llccc}
\toprule
\textbf{Task} & \textbf{Method} & \textbf{Valid runs} & \textbf{BIC Mean $\pm$ Std} & \textbf{BIC Range} \\
\midrule
Planning & DisCo & 5/5 & $221.32 \pm 2.83$ & [216.20, 224.94] \\
         & GeCCo & 3/5 & $231.65 \pm 16.58$ & [218.95, 255.06] \\
\midrule
Learning & DisCo & 5/5 & $74.49 \pm 2.20$ & [70.27, 76.71] \\
         & GeCCo & 3/5 & $80.57 \pm 7.18$ & [73.96, 90.55] \\
\midrule
Memory   & DisCo & 5/5 & $56.03 \pm 1.93$ & [53.46, 59.28] \\
         & GeCCo & 3/5 & $837.91 \pm 25.77$ & [802.73, 863.73] \\
\bottomrule
\end{tabular}
\end{table}

For all three tasks, DisCo produced a valid final model in each of the five runs, whereas GeCCo did so in only three of the five runs. 
Among the final models selected by GeCCo, some violated the Validator's constraints by making their predictions depend on observed behavioral responses, whereas others failed additional post-hoc validity checks because of incorrect likelihood computation. 
% likelihood computation 왜 오류 발생하는지 (disco, gecco 차이 - likelihood를 어디에서 계산/누가 정의하느냐 차이)
In GeCCo, the LLM can modify the code used to compute the likelihood, allowing errors in the likelihood computation (e.g., choice probabilities that do not sum to one) to persist in the final selected model. In DisCo, the LLM only generates choice probabilities, while the likelihood is computed separately by a fixed evaluation function. After excluding invalid runs, DisCo showed a lower mean BIC score than GeCCo for each task, with substantially lower variability across runs.

% held-out BIC 평가 방식이 어떻게 됐는지 안적혀잇음. gecco에게는 공정한 비교가 아닐 가능성.(을 너무 대놓고 말하지는 않지만 이야기는 해야함) -- 
The difference between the two methods was particularly pronounced in the memory task. 
The final DisCo models achieved held-out BIC scores of 53.46--59.28, whereas the valid final GeCCo models yielded scores of 802.73--863.73.
This difference may be partly explained by how candidate models are evaluated during the search.
DisCo splits each participant's data into training and validation sets for model evaluation during the search (see Appendix~\ref{app:dataset}), whereas GeCCo uses the full data without such a split. The final comparison applied the same splitting procedure used during DisCo’s search to a separate held-out dataset. DisCo’s training–validation split may help reduce overfitting, while also aligning its model-evaluation procedure more closely with the final held-out evaluation. Both factors may contribute to its lower BIC scores. However, the use of different data-splitting strategies does not affect the results of the validity checks, which show a key advantage of DisCo.
%This may help DisCo avoid selecting models that overfit the data, potentially contributing to its lower BIC scores in the final evaluation on held-out data.
%DisCo scores candidate models using a chronological validation split, whereas GeCCo scores them on the full response sequence.
% This evaluation scheme may reduce the likelihood of selecting models that fit the observed responses well but generalize poorly to unseen responses.
%evaluation 명확하게 하기
Additional details, including BIC scores under an alternative evaluation protocol, validation-rule violations in GeCCo-generated models, and qualitative comparisons between the models discovered by DisCo and GeCCo, are provided in Appendix~\ref{app:gecco}.

\subsection{Validator and Cleaner Ablation Study}
\label{sec:ablation}
To understand the role of each structural component in the DisCo pipeline, we conducted a $2 \times 2$ ablation study varying the presence of the Validator and Cleaner modules. Experiments were run on the planning task dataset, with 200 iterations per seed. To assess the utility of the modules, we quantified (i) the proportion of models violating the Validator's constraints across conditions and (ii) the BIC scores of the discovered models.

The results in Table \ref{tab:ablation} show that the Validator and Cleaner modules play complementary but distinct roles in the DisCo pipeline. The Cleaner primarily contributes to search stability: when present, the variance of BIC 
remains consistently low across seeds (MF: $1.75$, Hybrid: $2.83$), whereas removing it leads to higher variance and occasional pathological collapse (MF without Cleaner or Validator: $158.35 \pm 65.54$). This suggests that the Cleaner prevents the search from exploiting degenerate solutions that achieve low BIC through overfitting rather than genuine model improvement. 
% Revised (0505)
The Hybrid condition, being more structurally constrained, shows lower violation rates than the MF condition and yields relatively stable results even without the Validator. However, the elevated variance in the Cleaner-only condition ($195.09 \pm 33.26$) indicates that such stability is not guaranteed, supporting the role of the Validator as an additional safeguard.

% Revised (0505)
The Validator module, by contrast, enforces behavioral constraints that BIC alone cannot capture. Removing the Validator results in non-negligible rates of constraint violations (non-deterministic violations: up to $17.8\%$, independence violations: up to $2.5\%$). In most conditions without the Validator, BIC scores were relatively low but highly variable, suggesting that constraint violations may produce artificially favorable BIC scores.
Representative examples of these violations at the code level are provided in Appendix~\ref{app:violation}.

\begin{table}
    \centering
    \small
    \caption{Ablation study on the planning task across Validator and Cleaner module combinations. Results are reported across five independent runs, each with 200 iterations. Violation rates are computed only among accepted candidate models.}
    \begin{tabular}{c|ccrrrrcc}
    \toprule
    \textbf{Baseline} & \textbf{Cleaner} & \textbf{Validator} & \textbf{Non-det.} & \textbf{Indep.} & \textbf{Future} & \textbf{Neg-idx} & \textbf{BIC Mean $\pm$ Std} \\
    \midrule
    \multirow{4}{*}{MF} 
     & \checkmark & \checkmark & 0\% & 0\% & 0\% & 0\% & $221.16 \pm 1.75$  \\
     & \checkmark &            & 14.3\% & 2.3\% & 0\% & 0.1\% & $199.41 \pm 38.05$  \\
     &            & \checkmark & 0\% & 0\% & 0\% & 0\% & $218.07 \pm 3.22$  \\
     &            &            & 17.8\% & 2.5\% & 0\% & 0\% & $158.35 \pm 65.54$  \\
    \midrule
    \multirow{4}{*}{Hybrid} 
     & \checkmark & \checkmark & 0\% & 0\% & 0\% & 0\% & $221.32 \pm 2.83$  \\
     & \checkmark &            & 6.8\% & 0.9\% & 0\% & 0.1\% & $195.09 \pm 33.26$  \\
     &            & \checkmark & 0\% & 0\% & 0\% & 0\% & $222.34 \pm 4.13$  \\
     &            &            & 6.7\% & 2.1\% & 0\% & 0\% & $224.52 \pm 3.51$  \\
    \bottomrule
    \end{tabular}
    \label{tab:ablation}
    \vspace{2pt} \\
    {\footnotesize Non-det.: non-deterministic violation; Indep.: independence violation; Future: future peeking; Neg-idx: negative indexing.}
\end{table}

\subsection{External Validation with Cognitive Measures}
% Revised (0505) 
% Validation against external measures is a common approach for assessing model validity.
% Cognitive models often include parameters that relate to underlying psychological or cognitive processes. We therefore examined whether model-derived parameters showed consistent associations with external behavioral measures not used during fitting. The two-step task dataset includes participant-level external measures (see Section~\ref{sec:dataset_twostep}), which we used to assess whether the Validator module improves the consistency of associations between model parameters and external measures.

Cognitive models often include parameters that relate to underlying psychological or cognitive processes. We therefore examined whether model-derived parameters showed consistent associations with external measures of psychological traits. For the planning and learning datasets, we correlated fitted model parameters with available participant-level cognitive or behavioral measures and applied FDR correction for multiple comparisons. 
Because the two modules affected external associations differently across tasks, we report in the main text the task-specific contrasts in which each module showed the clearest effect, while providing the full results of FDR-significant associations in Appendix~\ref{app:external_validity}. 

%같은 과제에서 cleaner는 상대적으로 제한적/ 분명하게 보이지 않음. validator가 correlation에 더 영향이 큼. 메인 테이블에서는 제외. 

% As shown in Table~\ref{tab:external_validity}, models discovered with the Validator consistently identify a model-based weight $w$ that negatively correlates with age and positively correlates with IQ across all five runs, consistent with prior findings that model-based behavior is more prevalent in younger and higher-ability individuals \citep{daw2011model}. % cite another reference??
% Without the Validator, the consistency of these associations is compromised. Some runs yield models with additional parameters that render all associations non-significant. Other runs violate behavioral constraints and achieve lower BIC scores, with significant correlations driven by uninterpretable parameters ($\gamma$, $\alpha$) that show inconsistent signs across runs. These results suggest that predictive fit alone is insufficient to support meaningful parameter interpretability.

% cleaner의 효과는 미미했다. 보니까 Cleaner도 있는게 더 흐름은 좋아보이긴 합니다. Learning task와 planning task 에서 하나씩 나온걸 설명해보면 되겠습니다. 과제에 따라 각각 다른 모듈이 효과적일 수 있었다 라는 식으로요.

% 그러면 어제 보내드렸던 table을 유지하고...? 진행하면 될까요? 네, Table size를 축약할 수 있으면 더 좋고요. 시도해보겠습니다..!
Table~\ref{tab:external_validity} shows representative runs that illustrate how the Validator and Cleaner affect parameter interpretability. 
% In the planning task (A), models with the Validator consistently recover a model-based weight ($w$) that negatively correlates with age and positively correlates with IQ, consistent with prior findings \citep{eppinger2013goals, otto2013working, otto2015cognitive}; removing the Validator yields runs with lower BIC but non-significant associations or sign-inconsistent correlations on less interpretable parameters (e.g., $\gamma$). 
% In the planning task (A), models discovered with the Validator consistently identify a model-based weight ($w$) that negatively correlates with age and positively correlates with IQ, consistent with prior findings \citep{eppinger2013goals, otto2015cognitive}. They also show an association involving $\beta$, which may point to an additional individual-difference signal beyond the previously reported effects on model-based weighting.
% Without the Validator, some runs still show similar associations between $w$ and the external measures. However, some other runs with anomalously lower BIC show additional or shifted associations involving other parameters such as $\gamma$ and $\alpha$.
In the planning task (A), models discovered with the Validator consistently identify a model-based weight ($w$) that negatively correlates with age and positively correlates with IQ, consistent with prior findings \citep{eppinger2013goals, otto2015cognitive}. They also show a stable correlation between inverse temperature ($\beta$) and IQ, consistent with prior evidence linking IQ to decision noise \citep{ger2024using}. Although some associations vary across seeds, the central associations ($w$ with age and IQ, and $\beta$ with IQ) remain stable across runs (see Appendix~\ref{app:external_validity}). 
Without the Validator, some runs still show similar associations, while others show lower BIC yet exhibit additional or shifted associations involving parameters such as $\gamma$ and $\alpha$.
% In the learning task (B), retaining the Cleaner yields models with some significant external associations, whereas removing it produces models with substantially lower BIC but mostly non-significant associations. 
In the learning task (B), retaining the Cleaner yields some models with significant external associations involving $\beta$, whereas removing it produces substantially lower BIC but mostly non-significant or less clearly interpretable associations. 
These examples suggest that the Validator and Cleaner contribute to parameter interpretability in task-dependent ways.

\begin{table}[t]
\centering
\small
\setlength{\tabcolsep}{4pt}
\caption{External validity of model parameters under ablated conditions, shown for representative runs and associations that illustrate key contrasts.}
\label{tab:external_validity}
\begin{minipage}[t]{0.53\textwidth}
\centering
\textbf{(A) Validator effect}\\Planning task (Hybrid baseline), Cleaner ON
\vspace{2pt}

\begin{tabular}{lcl}
\toprule
\textbf{Condition} & \textbf{BIC} & \textbf{Significant correlations} \\
\midrule
w/ Validator 
  & 224.94 & $w \leftrightarrow \mathrm{age}\ {-}.19$, $\mathrm{IQ}\ {+}.19$ \\
  &        & $\beta \leftrightarrow \mathrm{IQ}\ {+}.17$ \\
  & 221.83 & $w \leftrightarrow \mathrm{age}\ {-}.22$, $\mathrm{IQ}\ {+}.12$ \\
  &        & $\beta \leftrightarrow \mathrm{IQ}\ {+}.15$ \\
  & 216.20 & $w \leftrightarrow \mathrm{age}\ {-}.23$, $\mathrm{IQ}\ {+}.16$ \\
  &        & $\beta \leftrightarrow \mathrm{IQ}\ {+}.14$ \\
\midrule
w/o Validator
  & 221.83 & $w \leftrightarrow \mathrm{age}\ {-}.22$, $\mathrm{IQ}\ {+}.12$ \\
  &        & $\beta \leftrightarrow \mathrm{IQ}\ {+}.15$ \\
  & 139.72 & $w \leftrightarrow \mathrm{age}\ {-}.15$, $\mathrm{IQ}\ {+}.12$ \\
  &        & $\gamma \leftrightarrow \mathrm{age}\ {+}.14$ \\
  & 173.32 & $\alpha \leftrightarrow \mathrm{IQ}\ {+}.20$, $\beta \leftrightarrow \mathrm{IQ}\ {-}.13$ \\
  &        & $\gamma \leftrightarrow \mathrm{age}\ {-}.20$ \\
\bottomrule
\end{tabular}
\end{minipage}
\hfill
\begin{minipage}[t]{0.45\textwidth}
\centering
\textbf{(B) Cleaner effect}\\
Learning task, Validator OFF
\vspace{3pt}

\begin{tabular}{lcl}
\toprule
\textbf{Condition} & \textbf{BIC} & \textbf{Significant correlations} \\
\midrule
w/ Cleaner
  & 76.06 & $\beta \leftrightarrow \mathrm{BIS}\ {-}.16$ \\
  & 76.06 & $\beta \leftrightarrow \mathrm{BIS}\ {-}.16$ \\
  & 75.37 & n.s. \\
\midrule
w/o Cleaner
  & 23.21 & n.s. \\
  & 29.79 & n.s. \\
  & 56.48 & $\gamma \leftrightarrow \mathrm{BIS}\ {-}.20$ \\
\bottomrule
\end{tabular}
\end{minipage}

\vspace{3pt}

{\footnotesize
All reported correlations are FDR-corrected $p < 0.05$. 
}
\end{table}

\subsection{Prompt Ablation Study}
% Revised (0505) 
% Motivated by prior LLM-based cognitive model discovery work that analyzed the contribution of prompt components through ablation \citep{rmus2025generating}, we similarly examined whether DisCo's prompt-level instructions affect the discovery of high-performing models. We conducted an ablation study over prompt-level components by systematically removing individual elements from the full DisCo pipeline and measuring the effect on the BIC scores of the discovered models. The Cleaner and Validator modules were retained across all conditions, as these constitute the core structural constraints of the framework. For this ablation study, we used the model-free (MF) formulation as the baseline model class, as its simpler parameterization provides a cleaner signal for isolating the effect of individual prompt components.
Motivated by prior LLM-based cognitive model discovery work that analyzed the contribution of prompt components through ablation \citep{rmus2025generating}, we examined whether DisCo's prompt-level instructions help guide the search toward high-performing models. Specifically, we systematically removed individual elements from the full DisCo prompt and measured their effects on the BIC scores of the discovered models. The Validator and Cleaner modules were retained across all conditions to isolate the effect of prompt-level guidance while preserving the core structural constraints. For this ablation study, we used the model-free (MF) formulation as the baseline model class, as its simpler parameterization provides a cleaner signal for comparing the effect of individual prompt components.

\begin{table}[t]
\centering
\small
\caption{Ablation study on prompt components. Each row removes one component from the full prompt. Lower BIC is better. Results are reported across five independent runs.}
\label{tab:ablation_prompt}
\begin{tabular}{lcc}
\toprule
\textbf{Prompt Configuration} & \textbf{BIC Mean $\pm$ Std} & \textbf{BIC range} \\
\midrule
Full prompt (all components)  & $221.16 \pm 1.75$ & [218.27, 223.57] \\
w/o Context                   & $220.29 \pm 3.93$ & [216.24, 227.46] \\
w/o Constraints               & $220.23 \pm 1.90$ & [218.25, 223.36] \\
w/o Parsimony                 & $224.61 \pm 3.50$ & [218.25, 228.17] \\
w/o Validation Method         & $219.77 \pm 2.03$ & [218.25, 223.36] \\
\bottomrule
\end{tabular}
\end{table}

Results shown in Table~\ref{tab:ablation_prompt} suggest that prompt-level optimization has limited effects on model quality, with BIC differences falling within run-to-run variability in most configurations. These results indicate that, in this setting, search outcomes are not strongly determined by any single prompt-level instruction, suggesting that prompt refinement alone may be insufficient for substantially improving the efficiency of model discovery.

\section{Conclusion}

We introduced DisCo, an LLM-guided framework for automatically discovering interpretable cognitive models from behavioral data. By combining LLM-based evolutionary search with structural constraints, DisCo aims to reduce the manual effort and theoretical bias inherent in traditional model construction. Across three cognitive tasks, DisCo discovered models with competitive or superior BIC scores compared to task-specific baselines, suggesting that LLM-guided search can offer a viable alternative to manual model construction.

However, the ablation study suggests that predictive performance alone is inadequate for evaluating model quality. Without the Validator, discovered models tend to violate constraints and show inconsistent associations with external cognitive measures, while occasionally achieving lower BIC scores. This dissociation between predictive fit and psychological validity suggests that effective LLM-guided model discovery requires integrating LLM-based search with domain-grounded validation.

Despite DisCo’s reliable performance, several limitations should be considered. First, its performance is sensitive to the provided baseline model, suggesting that prior assumptions derived from domain knowledge can shape the search process in both beneficial and detrimental ways. %Second, DisCo was evaluated on cognitive tasks assessing planning, learning, and memory. Whether the framework generalizes to other areas, such as perception, language processing, or motor control, remains to be tested.
Second, although DisCo automates model discovery, the Validator relies in part on task-specific constraints. %Extending DisCo to new domains may require appropriate baseline models and validation criteria. 
Establishing standardized evaluation protocols that extend beyond the validation criteria considered here, therefore remains an important open problem for cognitive model discovery. Looking forward, we hope DisCo contributes to a broader conversation about how cognitive modeling can be automated without compromising psychological validity and interpretability.

\begin{ack}
This work was supported by the New Faculty Settlement Research Program through Korea Advanced Institute of Science \& Technology (Grant No. G04240059 [to SHL]); the Data Science Convergence Human Resource Development Program through the National Research Foundation (NRF) of Korea funded by the Ministry of Science and ICT (Grant No. RS-2022-NR068758 [to SHL]); the AI Science Hub Program through the Ministry of Science and ICT (Grant No. N10260136 [to SHL]); and the Industry Research and Development Program funded by FMMENTALCARE (Grant No. G01250572 [to SHL]).
\end{ack}

\bibliography{example_paper}
\bibliographystyle{apalike}

% References follow the acknowledgments in the camera-ready paper. Use unnumbered first-level heading for
% the references. Any choice of citation style is acceptable as long as you are
% consistent. It is permissible to reduce the font size to \verb+small+ (9 point)
% when listing the references.
% Note that the Reference section does not count towards the page limit.
% \medskip

% {
% \small

% [1] Alexander, J.A.\ \& Mozer, M.C.\ (1995) Template-based algorithms for
% connectionist rule extraction. In G.\ Tesauro, D.S.\ Touretzky and T.K.\ Leen
% (eds.), {\it Advances in Neural Information Processing Systems 7},
% pp.\ 609--616. Cambridge, MA: MIT Press.

% [2] Bower, J.M.\ \& Beeman, D.\ (1995) {\it The Book of GENESIS: Exploring
%   Realistic Neural Models with the GEneral NEural SImulation System.}  New York:
% TELOS/Springer--Verlag.

% [3] Hasselmo, M.E., Schnell, E.\ \& Barkai, E.\ (1995) Dynamics of learning and
% recall at excitatory recurrent synapses and cholinergic modulation in rat
% hippocampal region CA3. {\it Journal of Neuroscience} {\bf 15}(7):5249-5262.
% }

%%%%%%%%%%%%%%%%%%%%%%%%%%%%%%%%%%%%%%%%%%%%%%%%%%%%%%%%%%%%
\newpage
\appendix

% \section{Technical appendices and supplementary material}
% Technical appendices with additional results, figures, graphs, and proofs may be submitted with the paper submission before the full submission deadline (see above). You can upload a ZIP file for videos or code, but do not upload a separate PDF file for the appendix. There is no page limit for the technical appendices. 

% Note: Think of the appendix as ``optional reading'' for reviewers. The paper must be able to stand alone without the appendix; for example, adding critical experiments that support the main claims to an appendix is inappropriate. 

%%%%%%%%%%%%%%%%%%%%%%%%%%%%%%%%%%%%%%%%%%%%%%%%%
\newpage
\section{Additional Dataset and Baseline Model Details}
\label{app:details}
\subsection{Baseline Model Specifications}
\subsubsection{Two-step Reinforcement Learning Task}
\label{app:twostep}

\paragraph{Model-Free Temporal-Difference Learning Model.}

The first baseline model is a simplified variant of the model-free temporal-difference (TD) learning model of \citet{daw2011model}. The model maintains separate action values for the first and second stages, updating first-stage values via credit assignment from second-stage outcomes without representing the task's transition structure.

The model maintains action values for the first stage, $Q_1(a)$, and state--action values for the second stage, $Q_2(s,a)$.
At each stage $i \in \{1,2\}$, choices are generated according to a softmax decision rule with inverse temperature parameter $\beta > 0$:
\[
  P(a_{i,t} = a \mid s_{i,t}) \propto \exp\!\bigl\{\beta\, Q_i(s_{i,t},a)\bigr\}.
\]
Learning proceeds sequentially within each trial.
Following the second-stage choice, the second-stage value is first updated using the obtained terminal reward $r_t$:
\[
  Q_2^{t+1}(s_{2,t},a_{2,t}) = Q_2^{t}(s_{2,t},a_{2,t}) + \alpha \bigl[r_t - Q_2^{t}(s_{2,t},a_{2,t})\bigr],
\]
where $\alpha \in [0,1]$ denotes the learning rate.
Subsequently, the first-stage action value is updated via credit assignment from the updated second-stage value estimate:
\[
  Q_1^{t+1}(a_{1,t}) = Q_1^{t}(a_{1,t}) + \alpha \bigl[Q_2^{t+1}(s_{2,t},a_{2,t}) - Q_1^{t}(a_{1,t})\bigr].
\]
Importantly, the first-stage update explicitly uses the post-update second-stage value, thereby implementing a fully model-free credit assignment mechanism without representing the task's transition structure.
Unlike the original formulation of \citet{daw2011model}, which includes separate learning rates and inverse temperatures for each stage, an eligibility trace parameter $\lambda$, and a perseveration parameter, the present implementation shares a single learning rate and inverse temperature across stages and omits the eligibility trace and perseveration terms, yielding a two-parameter model ($\alpha$, $\beta$) that prioritizes parsimony.

\paragraph{Hybrid Model-Based and Model-Free Learning Model.}

The second baseline integrates model-free (MF) temporal-difference learning with model-based (MB) planning using a mixing weight $w \in [0,1]$, following the hybrid architecture of \citet{daw2011model}. First-stage values are computed as a weighted combination of cached model-free values and model-based values obtained via one-step look-ahead using the known transition probabilities.
Both models use softmax action selection and TD learning for value updates.
This model maintains a model-free first-stage value $Q_{\text{MF}}(a_1)$, which is updated identically to $Q_1$ in the model-free baseline above, alongside second-stage values $Q_2(s,a)$.
At the first stage, the net action value is computed as a convex combination of MF and MB values:
\[
  Q_{\text{net}}(a_{1,t}) = w \cdot Q_{\text{MB},t}(a_{1,t}) + (1 - w) \cdot Q_{\text{MF},t}(a_{1,t}),
\]
where $w=0$ corresponds to purely model-free control and $w=1$ to purely model-based control.
The model-based value is obtained via a one-step look-ahead using the known transition structure of the task:
\[
  Q_{\text{MB},t}(a_{1,t}) = \sum_{s_2} P(s_2 \mid a_{1,t}) \max_{a_2} Q_{2,t}(s_2,a_2).
\]
First-stage choices are generated according to $P(a_{1,t} = a) \propto \exp\!\bigl\{\beta\, Q_{\text{net},t}(a)\bigr\}$, while second-stage choices follow $P(a_{2,t} = a \mid s_{2,t}) \propto \exp\!\bigl\{\beta\, Q_{2,t}(s_{2,t},a)\bigr\}$.
As with the model-free baseline, learning rates and inverse temperatures are shared across stages and the eligibility trace and perseveration parameters are omitted, resulting in a three-parameter model ($\alpha$, $\beta$, $w$).

\subsubsection{Probabilistic Reversal Learning Task: Temporal-Difference Learning Model}
\label{app:plt}

%\subsubsection{Temporal-Difference Learning Model}

The temporal-difference learning model \citep{mihatsch2002risk} maintains stimulus values $V(j)$ for each of the three options $j \in \{0,1,2\}$, initialized at zero.
On each trial $t$, choices are generated according to a softmax decision rule with inverse temperature $\beta > 0$:
\[
  P(c_t = j) \propto \exp\!\bigl\{\beta\, V_t(j)\bigr\}.
\]
Following each choice, a prediction error is computed as $\delta_t = o_t - V_t(c_t)$, where $o_t \in \{0,1\}$ denotes the observed outcome.
The chosen stimulus value is then updated asymmetrically depending on the sign of the prediction error:
\[
  V_{t+1}(c_t) = V_t(c_t) + \eta(\delta_t) \cdot \delta_t, \quad \text{where} \quad \eta(\delta_t) = \begin{cases} \eta_+ & \text{if } \delta_t \geq 0, \\ \eta_- & \text{if } \delta_t < 0, \end{cases}
\]
with $\eta_+, \eta_- \in [0,1]$ denoting separate learning rates for positive and negative prediction errors, respectively.
Values for unchosen stimuli remain unchanged.
This asymmetry allows the model to capture individual differences in the relative weighting of better- and worse-than-expected outcomes. When $\eta_- > \eta_+$, negative prediction errors have a larger impact on subsequent values than positive prediction errors, which can be interpreted as relatively greater sensitivity to losses or non-rewards; the reverse holds when $\eta_+ > \eta_-$.

\subsubsection{Recognition Memory Task: Signal Detection Model}
\label{app:memory}

%\subsubsection{Signal Detection Model}
We used a two-parameter equal-variance signal detection theory (SDT) model for the recognition memory task \citep{macmillan2005detection, stanislaw1999calculation}. Whereas the original study reported SDT measures by computing $d'$ and $c$ from hit and false-alarm rates, we implement the same equal-variance SDT assumptions as a two-parameter generative model for trial-level likelihood evaluation. The model assumes that each presented item elicits a continuous memory-strength signal drawn from one of two Gaussian distributions: $\mathcal{N}(d'/2,\,1)$ for old items and $\mathcal{N}(-d'/2,\,1)$ for new items, where $d'$ denotes discriminability.
The observer responds ``Yes'' whenever this signal exceeds a criterion $c$, yielding the following endorsement probabilities:
\[
  P(\text{Yes} \mid \text{old}) = \Phi\!\bigl(d'/2 - c\bigr), \qquad P(\text{Yes} \mid \text{new}) = \Phi\!\bigl(-d'/2 - c\bigr),
\]
where $\Phi(\cdot)$ denotes the standard normal cumulative distribution function.
A liberal criterion ($c < 0$) produces elevated endorsement rates for both old and new items, whereas a conservative criterion ($c > 0$) suppresses them.
The model has two free parameters: $d'$ and $c$.

\subsection{External Measures}
\label{app:external_measures}

In the main text, we reported external validity results for the planning and learning tasks. Here we describe the external measures available for each dataset.

\paragraph{Two-step reinforcement learning task.}
The dataset from \citet{gillan2016characterizing} included the following self-report measures collected alongside the behavioral task: the Obsessive-Compulsive Inventory--Revised (OCI-R), the Self-Rating Depression Scale (SDS), the trait portion of the State-Trait Anxiety Inventory (STAI), and an estimate of general cognitive ability (IQ). Demographic variables included participant age and gender.
 
\paragraph{Probabilistic reversal learning task.}
The dataset from \citet{zorowitz2023inattentive} included six self-report symptom and personality-trait measures administered before the behavioral task: the 7-Down scale, assessing depressive symptoms; the 7-Up scale, assessing hypomanic symptoms; the Generalized Anxiety Disorder 7-item scale (GAD-7); the Behavioral Inhibition System/Behavioral Activation System scales (BIS/BAS); the Snaith-Hamilton Pleasure Scale (SHAPS), assessing anhedonia; and the Penn State Worry Questionnaire (PSWQ). Demographic variables included participant age and gender.
 
\paragraph{Recognition memory task.}
The dataset from \citet{cataldo2023abnormal} included depressive symptom severity measured by the Beck Depression Inventory--II (BDI-II), along with participant age and gender. No additional clinical or personality measures were available in this dataset.

%%%%%%%%%%%%%%%%%%%%%%%%%%%%%%%%%%%%%%%%%%%%%%%%%%%
\newpage
\section{Dataset Splits and Model Evaluation}
\label{app:dataset}

\begin{figure}[H]
    \centering
    \includegraphics[width=\columnwidth]{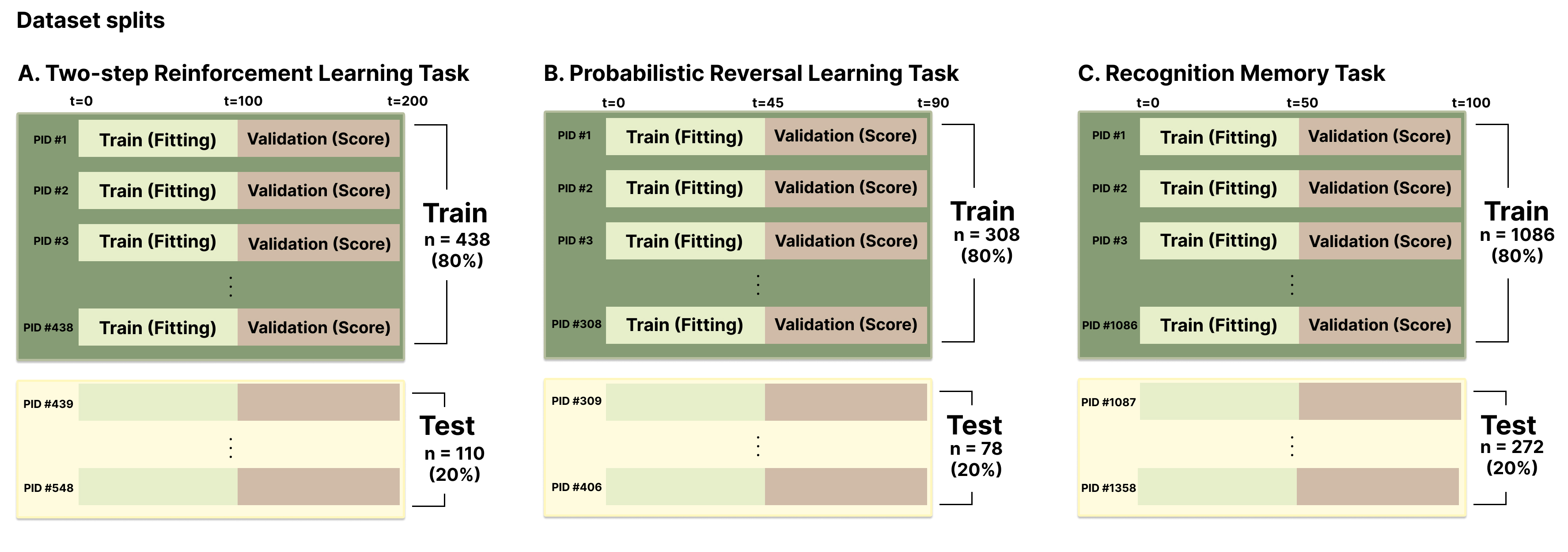}
    \caption{Overview of the behavioral datasets and data-splitting procedure. Participants were divided into train (80\%) and test (20\%) sets using random sampling. %random split 내용 추가
}
    \label{fig:dataset_split}
\end{figure}

To reduce overfitting during model selection and evaluate generalization to data from unseen participants, we employed a two-level data splitting strategy (Figure~\ref{fig:dataset_split}).
At the participant level, the full sample was divided into train (80\%) and test (20\%) sets using random sampling. The held-out test set was reserved exclusively for final model evaluation after all model selection procedures were completed using the train set. 

At the trial level, each participant's behavioral data were split temporally. For the two-step reinforcement learning task (200 trials), the first 100 trials were used for model fitting and the remaining 100 trials for computing validation scores. For the probabilistic reversal learning task (90 trials), the first 45 trials were used for fitting and the last 45 for validation. Lastly, for the recognition part of the memory task (100 trials), the first 50 trials were used for fitting and the last 50 trials for validation. During validation, model parameters remain fixed at the values estimated from the first half of the trials, while the model's latent states (e.g., Q-values) carry over and continue to update (see Optimization Objective in Section~\ref{sec:problem}).

We adopted this data-splitting strategy to assess generalization to unseen trials while preserving the temporal order of learning. This was particularly important for learning tasks, in which behavior in later trials (i.e., second half) may depend on learning experiences in early trials (i.e., first half). Reversing the training and validation halves, as in bidirectional split-half cross-validation, would be unsuitable for evaluating predictions of future behavior.

However, this splitting method may not be appropriate for all cognitive tasks, particularly when the two halves differ qualitatively or when splitting the trial sequence is impractical. 
Therefore, DisCo allows users to specify alternative evaluation strategies, including using the full trial sequence without splitting, n-fold cross-validation, and random trial-level splitting, depending on the characteristics of the task.

%%%%%%%%%%%%%%%%%%%%%%%%%%%%%%%%%%%%%%%%%%%%%%%%%
\newpage
\section{DisCo Algorithm}
\label{app:algorithm}
Algorithm~\ref{alg:pipeline} summarizes the full DisCo procedure, including LLM-based candidate generation, logical validation, validation-set scoring, and periodic pruning and migration across island buffers.

\begin{algorithm}
   \caption{DisCo: Discovery of Interpretable Symbolic Cognitive Models}
   \label{alg:pipeline}
\begin{algorithmic}[1]
   \Require Task $\mathcal{T}$, Template $\mathcal{S}$, behavioral data $\mathcal{D}$, optional external measures $\mathcal{M}$
   \Ensure Optimal cognitive model $\phi^*$
   \State $(\mathcal{D}_{\text{train}}, \mathcal{D}_{\text{val}}) \leftarrow \text{ChronologicalSplit}(\mathcal{D})$
   \State Initialize island buffers $\mathcal{B} = \{B_1, \ldots, B_K\}$ with seed programs from $\mathcal{S}$
   
   \While{budget not exhausted}
      \State \textcolor{gray}{// Evolutionary Hypothesis Generation}
      \State Select island $B_k$ and sample exemplars $\mathcal{P} \sim B_k$
      \State $\Phi_{\text{new}} \leftarrow \text{LLM.evolve}(\mathcal{T}, \mathcal{P})$
      
      \State \textcolor{gray}{// Model Validation}
      \ForAll{$\phi \in \Phi_{\text{new}}$}
         \If{$\text{LogicalValidation}(\phi)$} 
            % \State $\hat{\theta} \leftarrow \text{MLE}(\phi, \mathcal{D}_{\text{train}})$
            % \State $s \leftarrow \text{BIC}(\phi, \hat{\theta}, \mathcal{D}_{\text{val}})$ \textcolor{gray}{// Chronological Split}
            \State $\hat{\theta} \leftarrow \text{MLE}(\phi, \mathcal{D}_{\text{train}})$
            \State $s \leftarrow \text{BIC}(\phi, \hat{\theta}, \mathcal{D}_{\text{val}} \mid \mathcal{D}_{\text{train}})$ \textcolor{gray}{// Compute BIC without resetting latent states}
            \State Add $(\phi, s)$ to $B_k$
         \EndIf
      \EndFor
      
      \State \textcolor{gray}{// Model Selection \& Pruning}
      \If{$t \bmod T_{\text{clean}} = 0$}
         \State $\mathcal{B} \leftarrow \text{Cleaner}(\mathcal{B})$ \textcolor{gray}{// Prune redundant params}
         \State $\text{Migrate}(\mathcal{B})$ \textcolor{gray}{// Exchange models between islands}
      \EndIf
   \EndWhile
   \State $\phi^* \leftarrow \operatorname*{argmin}_{\phi \in \Phi_{\text{candidates}}} \text{BIC}(\phi)$
   \If{$\mathcal{M}$ is available}
      \State $\mathcal{R}_{\text{ext}} \leftarrow \text{Reporter}(\phi^*, \mathcal{M})$ \textcolor{gray}{// Summarize parameter--measure associations}
   \EndIf
   \State \textbf{return} $\phi^*$
   
\end{algorithmic}
\end{algorithm}

%BIC 부분은 아래에 설명 넣기

%%%%%%%%%%%%%%%%%%%%%%%%%%%%%%%%%%%%%%%%%%%%%%%%%
\newpage
\section{Prompts}
This section provides the task-specific candidate-generation prompts for each behavioral task, as well as the Cleaner prompt used for parameter pruning.
\label{app:prompt}
\subsection{Two-step Reinforcement Learning Task}
\label{app:twostep_prompt}
\begin{tcolorbox}[
  title={LLM Prompt for two-step reinforcement learning task},
  colback=white,
  colframe=black,
  boxrule=0.5pt,
  fontupper=\footnotesize
]
\begin{verbatim}
"""
Write a Python function for the Two-step Decision-Making Task.

[Context]
- Domain: Cognitive Science
- Goal: Maximize reward by making two consecutive choices.
- Structure: 
  Stage 1 action (0/1) determines Stage 2 state (0/1) via a FIXED transition matrix:
  * Action 0 -> State 0 (70%), State 1 (30%)
  * Action 1 -> State 1 (70%), State 0 (30%)

[Constraints]
1. No `np.clip`, `np.random`, or external libraries (except `numpy` as `np`).
2. Minimize parameter count (Parsimony).
3. Do not modify Input/Output signatures.
4. No peeking: probs[t] must not depend on reward[t] (only past rewards).
5. No action cheating: probs1[t] must not use a1[t], probs2[t] must not use a2[t].
6. No negative indexing at t=0: avoid a1[t-1] without checking t > 0.

[Evaluation]
- BIC on train/validation splits.
- Validate no peeking, no action cheating, no negative indexing.
- Validate probability structure.
- Max parameters <= 10. But, less is better (Because BIC penalizes number of params).

[Input]
Dictionary `data` with fields (all shape: (n_trials,)):
- 'stage1_response': int {0,1} (Choice 1)
- 'stage2_response': int {0,1} (Choice 2)
- 'stage2_state'   : int {0,1} (State reached)
- 'reward'         : int {0,1} (Outcome)

[Output]
Return `((probs1, probs2), bounds)`:
- probs1, probs2: (n_trials, 2) arrays, rows sum to 1.0.
- bounds: list of (min, max) for each param.
"""

@evaluator.run
{evaluator_function}

@equation.evolve
{equation_to_evolve}
\end{verbatim}
\end{tcolorbox}

\newpage
\subsection{Probabilistic Reversal Learning Task}
\label{app:plt_prompt}
\begin{tcolorbox}[
  breakable,
  enhanced,
  title={LLM Prompt for probabilistic reversal learning task},
  colback=white,
  colframe=black,
  boxrule=0.5pt,
  fontupper=\footnotesize
]
\begin{Verbatim}[breaklines=true, breakanywhere=true]
"""
Write a Python function for the Probabilistic Reward Learning (PRL) Task.

[Context]
- Domain: Cognitive Science
- Goal: Model how participants learn to choose the most rewarding option.
- Structure:
  6 blocks × 15 trials = 90 trials per participant.
  Each trial: participant selects one of 3 stimuli → receives binary reward (0 or 1).
  One stimulus is "correct" (higher reward probability) per block; it changes 
  every 15 trials (reversal).
  The model must learn option values from experience (choices and outcomes), 
  WITHOUT knowing  which stimulus is correct.

[Constraints]
1. No `np.clip`, `np.random`, or external libraries (except `numpy` as `np`).
2. Minimize parameter count (Parsimony).
3. Do not modify Input/Output signatures.
4. No peeking: probs[t] must not depend on outcomes[t] (only past outcomes).
5. No action cheating: probs[t] must not depend on choices[t] (only past choices).
6. No negative indexing at t=0: avoid outcomes[t-1] or choices[t-1] without 
checking t > 0.

[Evaluation]
- BIC on train/validation splits (50:50 temporal split within each participant).
- Validate no peeking, no action cheating, no negative indexing.
- Validate probability structure.
- Max parameters <= 10. But, less is better (Because BIC penalizes number of params).

[Input]
Dictionary `data` with fields:
- 'choices'   : (n_trials,) int {0, 1, 2} — chosen stimulus index each trial
- 'outcomes'  : (n_trials,) int {0, 1}    — reward received each trial
- 'n_options' : int (always 3)             — number of available stimuli

[Output]
Return `((probs,), bounds)`:
- probs  : (n_trials, n_options) array — rows must sum to 1.0, values in [0, 1].
- bounds : list of (min, max) for each param.
"""


@evaluate.run
{evaluator_function}

@equation.evolve
{equation_to_evolve}

\end{Verbatim}
\end{tcolorbox}

\newpage
\subsection{Memory Task}
\label{app:memory_prompt}
\begin{tcolorbox}[
  breakable,
  enhanced,
  title={LLM Prompt for recognition memory task},
  colback=white,
  colframe=black,
  boxrule=0.5pt,
  fontupper=\footnotesize
]
\begin{Verbatim}[breaklines=true, breakanywhere=true]
"""
Write a Python function for Recognition Memory Task.

[Context]
- Domain: Cognitive Science / Memory
- Goal: Model the probability of recognizing a word as "old" or "new".
- Structure:
  Each trial: Presented word (old/new) → Participant responds Yes/No

[Constraints]
1. No `np.clip`, `np.random`, or external libraries (except `numpy` as `np` and `scipy`).
2. Minimize parameter count (Parsimony).
3. Do not modify Input/Output signatures.
4. The function must NOT use participant responses — only `is_old` is available as input.

[Evaluation]
- BIC on train/validation splits (50:50 within-subject temporal split).
- Max parameters <= 10. But, less is better (Because BIC penalizes number of params).

[Input]
Dictionary `data` with fields (all shape: (n_trials,)):
- 'is_old' : bool — True if the word was previously studied (old), False if new

[Output]
Return `(p_yes, bounds)`:
- p_yes  : (n_trials,) array — P(response=1 | trial) for each trial
- bounds : list of (min, max) for each param
"""

@evaluate.run
{evaluator_function}

@equation.evolve
{equation_to_evolve}
\end{Verbatim}
\end{tcolorbox}

\newpage
\subsection{Cleaner Prompt}
\label{app:prompt_cleaner}

This prompt is used when the DisCo pipeline detects significant flaws in the generated model codes.
\begin{tcolorbox}[
  title={LLM Prompt for Cleaner},
  colback=white,
  colframe=black,
  boxrule=0.5pt,
  fontupper=\footnotesize
]
\begin{verbatim}
You are an expert in cognitive modeling parameter optimization.

**TASK:** Clean the following function by removing or merging redundant parameters.

**ORIGINAL FUNCTION:**
```python
{function_code}
```

**REQUIRED CHANGES (based on statistical analysis across subjects):**
{chr(10).join(f"{i+1}. {inst}" for i, inst in enumerate(instructions))}

**RULES:**
1. Remove or merge ONLY the parameters mentioned above
2. Update the bounds list accordingly (remove bounds for removed parameters)
3. Adjust parameter indexing 
   (e.g., if you remove params[1], params[2] becomes params[1])
4. Keep the core model logic intact
5. Ensure the function still produces valid probability outputs

**OUTPUT:**
Return ONLY the cleaned Python function code.
\end{verbatim}
\end{tcolorbox}
%%%%%%%%%%%%%%%%%%%%%%%%%%%%%%%%%%%%%%%%%%%%%%%%
\newpage
\section{Additional Configuration Details}
\label{app:config}
In addition to the settings reported in the main text, all DisCo experiments used 10 islands, generated four candidate programs per LLM prompt, included two exemplar functions as in-context examples per prompt, and used an island reset period of 3600 seconds. The Cleaner used thresholds of 0.9 for parameter correlation, 0.8 for covariance, and 0.01 for the variance ratio. External associations were assessed using Benjamini--Hochberg FDR correction with $p_{\mathrm{FDR}} < 0.05$, and candidate evaluation used a 60-second timeout per model. Candidate model parameters were estimated using constrained maximum likelihood estimation with L-BFGS-B.

Experiments were run on a workstation with an AMD Ryzen Threadripper PRO 5965WX CPU (24 cores / 48 threads), 512 GB RAM, and two NVIDIA RTX A6000 GPUs with 48 GB VRAM each. Each seed required approximately 1--2 hours of wall-clock time, depending on the task and the generated candidate models. Code and reproduction instructions are included in the supplementary materials.

\begin{table}[h]
\centering
\small
\caption{External assets used in this work.}
\label{tab:external_assets}
\begin{tabularx}{\textwidth}{p{0.15\textwidth} p{0.20\textwidth} p{0.20\textwidth} X}
\toprule
\textbf{Asset} & \textbf{Source} & \textbf{Use in this work} & \textbf{License / Usage information} \\
\midrule
Gillan et al. (2016) dataset 
& Public dataset / paper 
& Planning task evaluation 
& No explicit license found; used for academic research. \\

Zorowitz et al. (2023) dataset 
& Public dataset / paper 
& Learning task evaluation 
& MIT license. \\

Cataldo et al. (2023) dataset 
& Public dataset / paper 
& Memory task evaluation 
& No explicit license found; used for academic research. \\

Qwen2.5-32B 
& Model repository 
& LLM-based model generation 
& See original model license / terms. \\

vLLM 
& Open-source software 
& LLM serving 
& See original software license. \\
\bottomrule
\end{tabularx}
\end{table}

%%%%%%%%%%%%%%%%%%%%%%%%%%%%%%%%%%%%%%%%%%%%%%%%%%%%%%%%%%%%

\newpage
\section{Best Discovered Model}
\label{app:best_model}
We present the best discovered models, selected by the lowest validation BIC across five random seeds. DisCo improves upon established models in two ways: by generating interpretable extensions of existing mechanisms and by identifying theoretically distinct candidate accounts that replace central explanatory mechanisms. The form of improvement depends on the task and the seed model.

\subsection{Planning}
\label{sec:best_model_twostep}

For the planning task, we considered three baseline models. The null baseline predicts choices at chance level and contains no free parameters. The model-free TD learning model updates first-stage values based on second-stage outcomes. The hybrid model extends this model by additionally incorporating model-based planning through a mixing weight (w).

The models discovered from these baselines differed in how they modified the corresponding baseline accounts. The models discovered from the null and model-free baselines (\ref{app:best_model_null} and~\ref{app:best_model_mf}, respectively) largely recovered mechanisms already present in the baseline models. 
We therefore focus on the model discovered from the hybrid baseline~(\ref{app:best_model_hybrid}), which extends the baseline account with an additional interpretable mechanism.

This model preserves the mixture of model-based and model-free learning in the hybrid baseline. It adds two components: a global value-decay term and an outcome-dependent bias against repeating the previous first-stage action following an unrewarded trial. Importantly, this bias affects choice separately from the Q-learning  value update. 
The discovered model therefore introduces an explicit outcome-dependent switching mechanism while preserving the core model-based and model-free structure of the hybrid baseline.

\subsubsection{Null Baseline}
\label{app:best_model_null}
\begin{tcolorbox}[
  title={Best discovered model from null baseline},
  colback=white,
  colframe=black,
  boxrule=0.5pt,
  fontupper=\footnotesize,
  breakable
]
\begin{verbatim}
def equation(data, params):
    """
    Improved model using Q-learning with decay for past rewards and incorporating
    state transitions explicitly.
    """
    n = len(data["stage1_response"])
    
    # Initialize Q-values for both stages
    Q1 = np.zeros(2)  # Q-values for stage 1 actions
    Q2 = np.zeros(2)  # Q-values for stage 2 actions
    
    alpha = params[0]  # Learning rate
    beta = params[1]   # Temperature parameter for softmax
    gamma = params[2]  # Discount factor for future rewards
    delta = params[3]  # Decay factor for past rewards
    
    # Ensure parameters are within valid ranges
    bounds = [
        (0.0, 1.0),  # Learning rate between 0 and 1
        (0.0, 10.0),  # Temperature parameter between 0 and 10
        (0.0, 1.0),  # Discount factor between 0 and 1
        (0.0, 1.0)   # Decay factor between 0 and 1
    ]
    
    probs1 = np.zeros((n, 2))
    probs2 = np.zeros((n, 2))
    
    for t in range(n):
        # Compute probabilities for stage 1 action
        prob_1_0 = np.exp(beta * Q1[0]) / (np.exp(beta * Q1[0]) + np.exp(beta * Q1[1]))
        prob_1_1 = 1 - prob_1_0
        probs1[t] = [prob_1_0, prob_1_1]
        
        # Update Q-values for stage 1 based on chosen action and transition probabilities
        if data['stage1_response'][t] == 0:
            next_state = data['stage2_state'][t]
            reward = data['reward'][t]
            Q1[0] = (1 - alpha) * Q1[0] + alpha * (reward + gamma * Q2[next_state])
        else:
            next_state = data['stage2_state'][t]
            reward = data['reward'][t]
            Q1[1] = (1 - alpha) * Q1[1] + alpha * (reward + gamma * Q2[next_state])
        
        # Compute probabilities for stage 2 action
        prob_2_0 = np.exp(beta * Q2[0]) / (np.exp(beta * Q2[0]) + np.exp(beta * Q2[1]))
        prob_2_1 = 1 - prob_2_0
        probs2[t] = [prob_2_0, prob_2_1]
        
        # Update Q-values for stage 2 based on chosen action and reward
        if data['stage2_response'][t] == 0:
            reward = data['reward'][t]
            Q2[0] = (1 - alpha) * Q2[0] + alpha * reward
        else:
            reward = data['reward'][t]
            Q2[1] = (1 - alpha) * Q2[1] + alpha * reward
        
        # Apply decay to Q-values
        Q1 *= (1 - delta)
        Q2 *= (1 - delta)
    
    return (probs1, probs2), bounds
\end{verbatim}
\end{tcolorbox}

\subsubsection{Model-free Baseline}
\label{app:best_model_mf}
\begin{tcolorbox}[
  title={Best discovered model from model-free baseline},
  colback=white,
  colframe=black,
  boxrule=0.5pt,
  fontupper=\footnotesize,
  breakable
]
\begin{verbatim}
def equation(data, params):
    """
    Improved Python function for the Two-step Decision-Making Task.
    Stage 1 action (0/1) determines Stage 2 state (0/1) via a FIXED transition matrix:
    - Action 0 -> State 0 (70%), State 1 (30%)
    - Action 1 -> State 1 (70%), State 0 (30%)
    """
    a1 = data["stage1_response"]
    a2 = data["stage2_response"]
    r  = data["reward"]
    s2 = data["stage2_state"]
    n = len(a1)
    
    # Parameters
    alpha = params[0]      # Learning rate
    beta  = params[1]      # Inverse temperature
    gamma = params[2]      # Decay factor for past rewards
    
    # Parameter bounds
    bounds = [
        (0.0, 1.0),        # alpha
        (0.0, 20.0),       # beta
        (0.0, 1.0),        # gamma
    ]
    
    def softmax(beta, q):
        exps = np.exp(beta * (q - np.max(q)))
        return exps / np.sum(exps)
    
    # Initialize Q-values
    Q1 = np.zeros(2)          # Stage 1 Q-values (Action L, R)
    Q2 = np.zeros((2, 2))     # Stage 2 Q-values (State 0/1, Action L/R)
    
    # Initialize outputs
    probs1 = np.zeros((n, 2))
    probs2 = np.zeros((n, 2))
    
    for t in range(n):
        # Stage 1 (Choice)
        probs1[t] = softmax(beta, Q1)
    
        # Stage 2 (Choice)
        s = s2[t]
        probs2[t] = softmax(beta, Q2[s])
    
        # Learning
        a1_t = a1[t]  # Stage 1 action at trial t
        a2_t = a2[t]  # Stage 2 action at trial t
    
        # Stage 2 Update
        td2 = r[t] - Q2[s, a2_t]
        Q2[s, a2_t] += alpha * td2
    
        # Stage 1 Update
        td1 = Q2[s, a2_t] - Q1[a1_t]
        Q1[a1_t] += alpha * td1
    
        # Apply decay factor to past Q-values
        if t > 0:
            Q1 *= gamma
            Q2 *= gamma
            
    # Must return (probs1, probs2), bounds
    return (probs1, probs2), bounds
\end{verbatim}
\end{tcolorbox}

\subsubsection{Hybrid Baseline}
\label{app:best_model_hybrid}
\begin{tcolorbox}[
  title={Best discovered model from hybrid baseline},
  colback=white,
  colframe=black,
  boxrule=0.5pt,
  fontupper=\footnotesize,
  breakable
]
\begin{verbatim}
def equation(data, params):
    """Improved version of `equation_v1` with decay factor."""
    a1 = data["stage1_response"]
    a2 = data["stage2_response"]
    r  = data["reward"]
    s2 = data["stage2_state"]
    n = len(a1)
    
    # Parameters
    alpha = params[0] # Learning rate
    beta  = params[1] # Inverse temperature
    w     = params[2] # Mixing weight (0: Pure MF, 1: Pure MB)
    bias  = params[3] # Bias towards avoiding previously punished actions
    decay = params[4] # Decay factor for Q-values
    
    # Parameter bounds
    bounds = [
        (0.0, 1.0),   # alpha
        (0.0, 20.0),  # beta
        (0.0, 1.0),   # w
        (-1.0, 1.0),  # bias
        (0.0, 1.0)    # decay
    ]
    
    def softmax(beta, q):
        exps = np.exp(beta * (q - np.max(q)))
        return exps / np.sum(exps)
    
    # Initialize Q-values 
    Q_mf = np.zeros(2)           
    Q2   = np.zeros((2,2))       
    prev_a1 = 0
    prev_a2 = 0
    
    # Initialize outputs
    probs1 = np.zeros((n,2))
    probs2 = np.zeros((n,2))
    
    # Transition probabilities
    trans_probs = np.array([
        [0.7, 0.3], # Action 0 -> [State 0, State 1]
        [0.3, 0.7]  # Action 1 -> [State 0, State 1]
    ])
    
    for t in range(n):
        # Stage 1 Decision
        # Calculate Model-Based Values (Q_mb)
        max_Q2 = np.max(Q2, axis=1) # [Max val of State 0, Max val of State 1]
        Q_mb = np.dot(trans_probs, max_Q2) 
    
        # Combine MB and MF (Q_net)
        Q_net = w * Q_mb + (1 - w) * Q_mf
    
        # Add bias towards avoiding previously punished actions
        if t > 0:
            if r[t-1] == 0:
                Q_net[prev_a1] -= bias
    
        # Action Selection using Q_net
        probs1[t] = softmax(beta, Q_net)
    
        # Stage 2 Decision
        s = s2[t]
        probs2[t] = softmax(beta, Q2[s])
    
        # Learning 
        a1_t = a1[t]
        a2_t = a2[t]
    
        # Stage 2 Update (Standard TD)
        td2 = r[t] - Q2[s, a2_t]
        Q2[s, a2_t] += alpha * td2
    
        # Stage 1 Update (Model-Free only)
        td1 = Q2[s, a2_t] - Q_mf[a1_t]
        Q_mf[a1_t] += alpha * td1
    
        # Apply decay to all Q-values
        Q_mf *= (1 - decay)
        Q2 *= (1 - decay)
    
        # Update previous actions
        prev_a1 = a1_t
        prev_a2 = a2_t
    
    return (probs1, probs2), bounds
\end{verbatim}
\end{tcolorbox}

%%%%%%%%%%%%%%%%%%%%%%%%%%%%%%%%%%%%%%%%%%%%%%%%%%%%%%%%%%%%

\newpage
\subsection{Learning}
\label{app:best_model_learning}

Two baseline models were tested for the learning task: the null baseline, which assigns equal choice probability to each option, and the TD learning baseline (Appendix~\ref{app:plt}).

The model discovered from the null baseline (\ref{app:best_model_null_prl})  corresponds to a restricted case of the TD learning baseline in which \(\eta_+ = \eta_-\), without a distinct learning mechanism.
In contrast, the model discovered from the TD learning baseline (\ref{app:best_model_td}) provides an interpretable alternative to the baseline. The model replaces the separate learning rates for positive and negative prediction errors with a single learning rate. It also introduces decay of unchosen-option values over time. As a result, individual differences are explained not by differential sensitivity to positive and negative outcomes, but by how quickly information about unchosen options is forgotten. A novel behavioral prediction of this account is that participants should switch more slowly to the newly advantageous option after a reversal when it has remained unchosen for longer.

\subsubsection{Null Baseline}
\label{app:best_model_null_prl}
\begin{tcolorbox}[
  title={Best discovered model from null baseline},
  colback=white,
  colframe=black,
  boxrule=0.5pt,
  fontupper=\footnotesize,
  breakable
]
\begin{verbatim}
def equation(data, params):
    """Improved version of `equation_v1` with a temperature parameter."""
    n_trials = len(data['choices'])
    n_options = data['n_options']
    
    # Initialize value estimates for each option
    values = np.zeros(n_options)
    
    # Learning rate (free parameter)
    alpha = params[0]
    
    # Temperature parameter (free parameter) to control exploration
    temperature = params[1]
    
    # Initialize probability matrix
    probs = np.zeros((n_trials, n_options))
    
    for t in range(n_trials):
        # Softmax transformation to get probabilities
        exp_values = np.exp(values / temperature)
        total_exp_value = np.sum(exp_values)
        probs[t] = exp_values / total_exp_value
        
        # Update values based on the outcome
        if t < n_trials - 1:
            chosen_option = data['choices'][t]
            reward = data['outcomes'][t]
            prediction_error = reward - values[chosen_option]
            values[chosen_option] += alpha * prediction_error
    
    # Bounds for the learning rate (alpha) and temperature
    bounds = [(0, 1), (0.1, 10)]
    
    return (probs,), bounds
\end{verbatim}
\end{tcolorbox}

\subsubsection{TD Learning Baseline}
\label{app:best_model_td}
\begin{tcolorbox}[
  title={Best discovered model from TD learning baseline},
  colback=white,
  colframe=black,
  boxrule=0.5pt,
  fontupper=\footnotesize,
  breakable
]
\begin{verbatim}
def equation(data, params):
    """
    Risk-Sensitive TD Learning with Decay for Probabilistic Reward Learning.
    
    params: [alpha, beta, gamma]
      alpha : learning rate (single rate for both positive and negative prediction errors)
      beta  : inverse temperature (softmax)
      gamma : decay factor to reduce the value of unchosen options over time
    """
    alpha, beta, gamma = params[0], params[1], params[2]
    bounds = [
        (0.0,  1.0),   # alpha
        (0.0, 20.0),   # beta
        (0.0,  1.0),   # gamma
    ]
    
    choices   = data['choices']
    outcomes  = data['outcomes']
    n_options = data['n_options']
    n_trials  = len(choices)
    
    V     = np.zeros(n_options)
    probs = np.zeros((n_trials, n_options))
    
    for t in range(n_trials):
        logits   = beta * (V - V.max())
        probs[t] = np.exp(logits) / np.exp(logits).sum()
    
        delta = outcomes[t] - V[choices[t]]
        V[choices[t]] += alpha * delta
    
        # Decay the value of unchosen options
        for i in range(n_options):
            if i != choices[t]:
                V[i] *= (1 - gamma)
    
    return (probs,), bounds
\end{verbatim}
\end{tcolorbox}

\newpage
\subsection{Memory}
\label{app:best_model_memory}

For the memory task, we considered two baseline models. The null baseline predicts ``yes'' at a constant rate. The SDT baseline (Appendix~\ref{app:memory}) predicts the responses using two parameters: the discriminability parameter $d'$ captures the separation between the memory strength distributions for old and new items, while the criterion $c$ sets the response threshold for a ``yes'' response.

The model discovered from the null baseline recovered separate parameters for the old-new distinction and overall response bias. The structure is analogous to the discriminability and criterion structure of the SDT baseline, although the discovered model uses a logit rather than the probit link used by the baseline. 

The model discovered from the SDT baseline preserves the discriminability and criterion structure but applies a bounded sigmoid transformation to the criterion $c$. In the standard SDT model, changes in $c$ continue to shift the decision threshold without saturation. In the discovered model, by contrast, the sigmoid transformation limits the influence of $c$ on the threshold at extreme values.

\subsubsection{Null Baseline}
\label{app:best_model_null_rm}
\begin{tcolorbox}[
  title={Best discovered model from null baseline},
  colback=white,
  colframe=black,
  boxrule=0.5pt,
  fontupper=\footnotesize,
  breakable
]
\begin{verbatim}
def equation(data, params):
    """
    Enhanced model: Introduces two parameters to capture the difference in recognition
    probabilities and overall response bias.
    - params[0]: bias (difference in log odds between old and new words)
    - params[1]: overall response bias (log odds of saying "Yes" regardless of old/new)
    """
    n = len(data['is_old'])
    bias = params[0]
    response_bias = params[1]
    
    # Compute the log odds for each trial
    log_odds = np.where(data['is_old'], bias + response_bias, response_bias)
    
    # Convert log odds to probabilities
    p_yes = 1 / (1 + np.exp(-log_odds))
    
    # Define bounds for the parameters
    bounds = [(-5, 5), (-5, 5)]
    
    return p_yes, bounds
\end{verbatim}
\end{tcolorbox}
\subsubsection{SDT Baseline}
\label{app:best_model_sdt}
\begin{tcolorbox}[
  title={Best discovered model from SDT baseline},
  colback=white,
  colframe=black,
  boxrule=0.5pt,
  fontupper=\footnotesize,
  breakable
]
\begin{verbatim}
def equation(data, params):
    """
    Improved version of `equation_v1` with a non-linear transformation of the criterion.
    
    Hit  = P("Yes" | old) = Phi( d_prime / 2 - sigmoid(c))
    FA   = P("Yes" | new) = Phi(-d_prime / 2 - sigmoid(c))
    
    Parameters:
    - d_prime: Sensitivity to the stimulus
    - c: Criterion (decision threshold)
    """
    d_prime, c = params[0], params[1]
    is_old = data['is_old']
    
    # Sigmoid function to transform the criterion
    sigmoid_c = 1 / (1 + np.exp(-c))
    
    p_hit = norm.cdf(d_prime / 2 - sigmoid_c)
    p_fa = norm.cdf(-d_prime / 2 - sigmoid_c)
    
    p_yes = np.where(is_old, p_hit, p_fa)
    
    bounds = [
        (-1.0, 10.0),  # d_prime
        (-3.0, 3.0),   # c
    ]
    
    return p_yes, bounds
\end{verbatim}
\end{tcolorbox}

%%%%%%%%%%%%%%%%%%%%%%%%%%%%%%%%%%%%%%%%%%%%%%%%%%%%%%%%%%%%

\newpage
\section{Comparison with GeCCo}
\label{app:gecco}
% 다시 확인해볼것 **
We provide additional details on the comparison between DisCo and GeCCo \citep{rmus2025generating} reported in Section~\ref{sec:gecco_comparison}. We first describe the experimental setup, followed by an analysis of validation-rule violations in GeCCo-generated models and a qualitative comparison of the final model structures discovered by the two methods.

\subsection{Experimental Setup}

We compared DisCo and GeCCo on the planning, learning, and memory tasks. For each task, both methods were initialized from the same task-specific baseline model: the hybrid model for planning, the TD learning model for learning, and the signal detection theory (SDT) model for memory. Each method was run with five independent seeds under the same computational budget and LLM model (Qwen2.5-32B), generating 200 candidate models per seed. The final selected model from each run was evaluated on held-out data using the same evaluation procedure described in the main text.

For the BIC comparison, we report statistics only for runs in which the final selected model satisfied the validation criteria. This prevents mathematically or structurally invalid models from being treated as competitive simply because they achieve favorable numerical scores. In particular, because GeCCo allows LLM-generated models to implement their own likelihood computation, the validation procedure was extended to check consistency between the generated probability outputs and the corresponding likelihood calculation.

\subsection{Validation-Rule Violations}

We additionally examined the candidate models generated by GeCCo for violations of the validation criteria used in our evaluation. Table~\ref{tab:gecco_violations} summarizes the frequency of each violation type across tasks.

\begin{table}[ht]
\centering
\small
\caption{Frequency of validation-rule violations among GeCCo-generated candidate models. A single candidate may violate more than one rule.}
\label{tab:gecco_violations}
\begin{tabular}{lccccc}
\toprule
\textbf{Task} &
\textbf{Future} &
\textbf{Independence} &
\textbf{Negative} &
\textbf{Non-} &
\textbf{Log-likelihood} \\
&
\textbf{peeking} &
\textbf{violation} &
\textbf{indexing} &
\textbf{determinism} &
\textbf{mismatch} \\
\midrule
Planning & 56 & 105 & 0 & 33 & 42 \\
Learning & 7  & 149 & 0 & 46 & 11 \\
Memory   & 0  & 69  & 1 & 83 & 51 \\
\bottomrule
\end{tabular}
\end{table}

Multiple types of violations occurred across the three tasks. Independence violations were particularly frequent in the planning and learning tasks, whereas uncontrolled non-determinism was common in the memory task. Future peeking and inconsistencies in likelihood computation were also observed, while negative indexing occurred only rarely.

Importantly, some of these violations affected final model selection. Across all three tasks, DisCo selected valid models in all five runs, whereas only three of five GeCCo runs produced valid final models. These results indicate that high model-fit scores alone do not necessarily imply that an LLM-generated model satisfies the structural and mathematical requirements of the task.

\subsection{Qualitative Comparison of Discovered Models}

Beyond validity and predictive performance, we qualitatively compared the structures of the final models selected by DisCo and GeCCo for each task.

\paragraph{Planning.}
The model discovered by DisCo retained the hybrid model-based/model-free architecture while introducing an outcome-dependent bias on the previous first-stage action. 
% Specifically, the bias promoted switching following nonreward, beyond changes explained by accumulated action values.
Specifically, the bias lowered the value of the previous first-stage action after an unrewarded trial. In contrast, the model discovered by GeCCo removed the explicit model-based/model-free mixture weight and did not include a mechanism that directly captured the effect of the previous outcome. Thus, the two methods produced structurally different accounts of behavior in the planning task.

\paragraph{Learning.} 
The models discovered by DisCo and GeCCo for the learning task were somewhat similar to each other. Both replaced the baseline model's separate learning rates for positive and negative prediction errors with a single learning rate and introduced a decay mechanism. However, DisCo applied decay selectively to unchosen-option values, whereas GeCCo applied decay more broadly to all option values. Both suggested that forgetting is a relevant mechanism, despite differences in how it is implemented. 

\paragraph{Memory.}
The DisCo model used a static, sigmoid-bounded decision criterion that depended on the stimulus but not on the participant's preceding responses. In contrast, the GeCCo models updated their decision criterion across trials based on response history. This allowed them to closely track the response sequence used for parameter estimation but led to poor generalization to held-out trials, as reported in the main text.

\subsection{Held-Out Evaluation with Full Sequence Fitting}

For the comparison in Table~\ref{tab:gecco_comparison}, we computed BIC scores for both methods using the data-splitting procedure that DisCo uses during model search. This alignment may favor DisCo. We therefore reevaluated the same set of valid final models under a protocol that matches how GeCCo fits models during search. Specifically, for each held-out test participant, we used the full trial sequence to estimate model parameters and to compute BIC.

%To further compare DisCo and GeCCo, we evaluated their final selected models under a full sequence fitting protocol (Table~\ref{tab:full_bic_comparison}). DisCo employs a split-half protocol, estimating model parameters on the first half of the trial sequence and evaluating BIC on the second half. During the second half, model parameters remain fixed, while latent states continue to be updated without resetting. In contrast, GeCCo uses the full trial sequence for parameter estimation. We therefore additionally evaluated both methods using GeCCo's fitting protocol, while retaining held-out data for the final evaluation.
\begin{table}[ht]
\centering
\small
\caption{Held-out BIC comparison between baseline models, DisCo, and GeCCo after fitting model parameters on the full trial sequence. Statistics are reported as mean $\pm$ standard deviation over valid runs, excluding runs in which the final selected model failed to satisfy the validation criteria.}
\label{tab:full_bic_comparison}
\begin{tabular}{llccc}
\toprule
\textbf{Task} & \textbf{Method} & \textbf{Valid runs} & \textbf{BIC Mean $\pm$ Std} & \textbf{BIC Range} \\
\midrule
Planning & Baseline (Hybrid) & -- & $478.05$ & -- \\
         & DisCo & 5/5 & $384.91 \pm 6.71$ & [372.55, 393.06] \\
         & GeCCo & 3/5 & $384.57 \pm 6.00$ & [377.50, 392.16] \\
\midrule
Learning & Baseline (TD Learning) & -- & $134.08$ & -- \\
         & DisCo & 5/5 & $125.02 \pm 2.67$ & [123.29, 130.31] \\
         & GeCCo & 3/5 & $125.41 \pm 1.76$ & [123.92, 127.88] \\
\midrule
Memory   & Baseline (SDT) & -- & $83.51$ & -- \\
         & DisCo & 5/5 & $89.92 \pm 4.80$ & [83.75, 95.85] \\
         & GeCCo & 3/5 & $76.87 \pm 0.39$ & [76.49, 77.41] \\
\bottomrule
\end{tabular}
\end{table}

Table~\ref{tab:full_bic_comparison} summarizes the results. Under the full-sequence evaluation protocol, which may favor GeCCo, DisCo and GeCCo achieved comparable BIC scores for held-out participants in the planning and learning tasks. In the planning task, DisCo obtained a mean BIC of 384.91, compared with 384.57 for GeCCo. Similarly, in the learning task, the mean BIC scores were 125.02 and 125.41, respectively. Note that the BIC scores are not directly comparable between Tables~\ref{tab:gecco_comparison} and \ref{tab:full_bic_comparison} because the log-likelihood term in BIC is summed over different numbers of trials in the two analyses.

%Despite their comparable performance, DisCo produced valid final models in all five runs for both tasks, whereas GeCCo produced valid models in only three out of five runs.

In the memory task, GeCCo achieved a lower mean BIC than DisCo (76.87 vs. 89.92). This contrasts with the primary analysis (Table~\ref{tab:gecco_comparison}), which used DisCo’s split-half protocol for BIC calculation.
%However, the opposite pattern was observed under DisCo's split-half protocol, where GeCCo obtained a substantially higher BIC than DisCo (837.91 vs. 56.03; Table~\ref{tab:gecco_comparison}).
This reversal suggests that the memory task may differ from planning and learning tasks in how consistently response patterns carry over from the first half to the second. Under the full-sequence protocol, shifts in response patterns across halves may be less apparent because BIC is computed on the same trials used to estimate model parameters. This protocol may favor more flexible models that can describe shifts in response patterns across the trial sequence, such as the history-dependent model discovered by GeCCo.
%This reversal may reflect differences in how well the models fit the estimation trials. Under the full-sequence protocol, parameters are estimated and evaluated on the same trials. which may favor models with more flexible structures, such as the GeCCo models with history-dependent decision criteria.
Under the split-half protocol, however, this flexibility may have led to overfitting, as suggested by the substantially higher BIC of GeCCo's valid final models (837.91 vs. 56.03 for DisCo). %Similarly, when fitted to only the first half of trials, the baseline SDT model can produce extreme response probabilities that generalize poorly to the second half. In contrast, DisCo's bounded decision criterion may reduce its sensitivity to extreme response patterns in the estimation trials, which may support better generalization to held-out trials.
DisCo appeared more robust across protocols: it showed a much smaller disadvantage under full-sequence evaluation than GeCCo did under split-half evaluation.

%Although the absolute BIC values across the two protocols are not directly comparable due to differences in the evaluated trial sequences, the reversal in relative performance suggests that the models' performance is sensitive to the fitting and evaluation protocol. Notably, even under the full sequence fitting protocol where GeCCo achieved a lower BIC in the memory task, it produced valid final models in only three out of five runs.
%its disadvantage under full-sequence fitting was modest, whereas GeCCo yielded substantially higher BIC scores under the split-half protocol.

%%%%%%%%%%%%%%%%%%%%%%%%%%%%%%%%%%%%%%%%%%%%%%%%%%%%%%%%%%%%

\newpage
\section{Representative Validator Violations}
\label{app:violation}

This appendix provides representative examples of validity violations detected by the Validator. While the main results in Section~\ref{sec:ablation} summarize violation rates across module ablations, here we illustrate what these violations look like at the program level. 
The examples are excerpts from generated candidate models, selected to illustrate the core invalid behavior.
Future-peeking violations were not observed among the accepted candidate models in our ablation study, so we do not include a separate example for this category.

\subsection{Non-determinism}
This example shows a model that uses a random draw while computing choice probabilities. 
Here, \texttt{epsilon} denotes an exploration probability, and \texttt{probs1[t]} denotes the predicted probability distribution over the two first-stage actions at trial $t$. 
Because \texttt{np.random.rand()} changes across repeated calls, the same input data and parameters can produce different predicted probabilities. 
\begin{tcolorbox}[
  title={},
  colback=white,
  colframe=black,
  boxrule=0.5pt,
  fontupper=\footnotesize,
  breakable
]
\begin{verbatim}
if np.random.rand() < epsilon:
    probs1[t] = np.ones(2) / 2
\end{verbatim}
\end{tcolorbox}

\subsection{Independence of Action}
This example shows a model that uses the observed current action while computing the probability of that same action. 
Here, \texttt{a1[t]} is the participant's observed first-stage action at trial $t$, \texttt{Q\_net} is the action-value vector used for prediction, and \texttt{probs1[t]} is the predicted choice probability. 
Because the model modifies \texttt{Q\_net} using \texttt{a1[t]} before computing \texttt{probs1[t]}, the predicted probability depends on the answer it is supposed to predict.
\begin{tcolorbox}[
  title={},
  colback=white,
  colframe=black,
  boxrule=0.5pt,
  fontupper=\footnotesize,
  breakable
]
\begin{verbatim}
# Add optimism bias
Q_net[a1[t]] += optim
        
# Action Selection using Q_net
probs1[t] = softmax(beta, Q_net)
\end{verbatim}
\end{tcolorbox}

\subsection{Negative Indexing}
This example shows a model that accesses the previous action using \texttt{a1[t-1]}. 
Here, \texttt{a1} is the sequence of observed first-stage actions and $t$ is the trial index. 
At the first trial ($t=0$), Python interprets \texttt{a1[t-1]} as \texttt{a1[-1]}, which refers to the final trial rather than a previous trial. 
This can inadvertently leak information from the end of the sequence into the first prediction.
\begin{tcolorbox}[
  title={},
  colback=white,
  colframe=black,
  boxrule=0.5pt,
  fontupper=\footnotesize,
  breakable
]
\begin{verbatim}
h1 = H1 + np.array([eta * (a1[t-1] == 0), eta * (a1[t-1] == 1)])
\end{verbatim}
\end{tcolorbox}

%%%%%%%%%%%%%%%%%%%%%%%%%%%%%%%%%%%%%%%%%%%%%%%%%%%%%%%%%%%%
\newpage
\section{External Validity Study}
\label{app:external_validity}
\begin{footnotesize}
\setlength{\tabcolsep}{4pt}

\begin{longtable}{llccrrp{0.45\textwidth}}
\label{tab:external_validity_full}\\  % \\ 도 빠져있음!
\caption{Full external validation results across random seeds. \checkmark\ indicates the module was active; $\times$ indicates it was disabled.} \\
\toprule
\textbf{Task} & \textbf{Model} & \textbf{Valid.} & \textbf{Clean.} & \textbf{Seed} & \textbf{BIC} & \textbf{FDR-significant correlations} \\
\midrule
\endfirsthead

\multicolumn{7}{c}{\textit{(Continued from previous page)}}\\
\toprule
\textbf{Task} & \textbf{Model} & \textbf{Valid.} & \textbf{Clean.} & \textbf{Seed} & \textbf{BIC} & \textbf{FDR-significant correlations} \\
\midrule
\endhead

\bottomrule
\endfoot

%% ── Planning / MF ─────────────────────────────────────────────
\multirow{20}{*}{Planning} & \multirow{20}{*}{MF}
  & \multirow{5}{*}{\checkmark} & \multirow{5}{*}{\checkmark}
  & 42   & 218.27 & $\alpha$: age ${+}.12$;\ $\beta$: IQ ${+}.16$ \\
& & & & 123  & 221.24 & $\beta$: IQ ${+}.16$ \\
& & & & 456  & 223.57 & $\alpha$: age ${+}.13$;\ $\beta$: IQ ${+}.16$;\ $\delta$: OCI ${+}.12$ \\
& & & & 789  & 222.09 & $\beta$: IQ ${+}.13$;\ $\gamma$: age ${+}.19$ \\
& & & & 1234 & 220.66 & $\alpha$: age ${+}.12$;\ $\beta$: IQ ${+}.17$ \\
\cmidrule(l){3-7}
& & \multirow{5}{*}{$\times$} & \multirow{5}{*}{\checkmark}
  & 42   & 123.48 & n.s. \\
& & & & 123  & 218.27 & $\alpha$: age ${+}.12$;\ $\beta$: IQ ${+}.16$ \\
& & & & 456  & 214.43 & $\alpha$: IQ ${+}.18$;\ $\beta$: age ${+}.13$;\ $\gamma$: age ${+}.17$ \\
& & & & 789  & 222.63 & $\alpha$: age ${+}.14$;\ $\beta$: IQ ${+}.15$;\ $\delta$: OCI ${+}.14$ \\
& & & & 1234 & 218.25 & $\alpha$: age ${+}.12$;\ $\beta$: IQ ${+}.15$;\ $\gamma$: OCI ${+}.12$ \\
\cmidrule(l){3-7}
& & \multirow{5}{*}{\checkmark} & \multirow{5}{*}{$\times$}
  & 42   & 212.72 & $\alpha$: age ${+}.14$;\ $\gamma$: age ${+}.13$, OCI ${-}.12$;\ $\delta$: OCI ${+}.16$ \\
& & & & 123  & 218.25 & $\alpha$: age ${+}.12$;\ $\beta$: IQ ${+}.15$;\ $\gamma$: OCI ${+}.12$ \\
& & & & 456  & 218.27 & $\alpha$: age ${+}.12$;\ $\beta$: IQ ${+}.16$ \\
& & & & 789  & 222.87 & $\alpha$: age ${+}.12$;\ $\beta$: IQ ${+}.16$ \\
& & & & 1234 & 218.25 & $\alpha$: age ${+}.12$;\ $\beta$: IQ ${+}.15$;\ $\gamma$: OCI ${+}.12$ \\
\cmidrule(l){3-7}
& & \multirow{5}{*}{$\times$} & \multirow{5}{*}{$\times$}
  & 42   & 208.13 & $\alpha$: IQ ${+}.15$;\ $\beta$: IQ ${-}.16$ \\
& & & & 123  & 36.76  & n.s. \\
& & & & 456  & 210.39 & $\beta$: IQ ${+}.15$ \\
& & & & 789  & 193.81 & n.s. \\
& & & & 1234 & 142.65 & $\alpha_1$: age ${+}.15$, IQ ${-}.16$;\ decay: IQ ${+}.14$ \\
\midrule

%% ── Planning / Hybrid ─────────────────────────────────────────
\multirow{20}{*}{Planning} & \multirow{20}{*}{Hybrid}
  & \multirow{5}{*}{\checkmark} & \multirow{5}{*}{\checkmark}
  & 42   & 224.94 & $\beta$: IQ ${+}.17$;\ $w$: age ${-}.19$, IQ ${+}.19$ \\
& & & & 123  & 221.83 & $\beta$: IQ ${+}.15$;\ $w$: age ${-}.22$, IQ ${+}.12$;\ $\gamma$: OCI ${-}.12$ \\
& & & & 456  & 221.83 & $\beta$: IQ ${+}.15$;\ $w$: age ${-}.22$, IQ ${+}.12$;\ $\gamma$: OCI ${+}.12$ \\
& & & & 789  & 221.83 & $\beta$: IQ ${+}.15$;\ $w$: age ${-}.22$, IQ ${+}.12$;\ $\gamma$: OCI ${-}.12$ \\
& & & & 1234 & 216.20 & $\beta$: IQ ${+}.14$;\ $w$: age ${-}.23$, IQ ${+}.16$;\ decay: OCI ${+}.15$ \\
\cmidrule(l){3-7}
& & \multirow{5}{*}{$\times$} & \multirow{5}{*}{\checkmark}
  & 42   & 221.83 & $\beta$: IQ ${+}.15$;\ $w$: age ${-}.22$, IQ ${+}.12$;\ $\gamma$: OCI ${+}.12$ \\
& & & & 123  & 221.83 & $\beta$: IQ ${+}.15$;\ $\gamma$: OCI ${+}.12$;\ $w$: age ${-}.22$, IQ ${+}.12$ \\
& & & & 456  & 218.75 & $\beta$: OCI ${+}.11$;\ $w$: age ${-}.14$, IQ ${+}.15$;\ $\gamma$: OCI ${+}.12$ \\
& & & & 789  & 139.72 & $\alpha$: IQ ${+}.11$;\ $\beta$: OCI ${+}.13$;\ $w$: age ${-}.15$, IQ ${+}.12$;\ $\gamma$: age ${+}.14$, OCI ${-}.11$;\ $\eta$: OCI ${+}.13$ \\
& & & & 1234 & 173.32 & $\alpha$: IQ ${+}.20$;\ $\beta$: IQ ${-}.13$;\ $\gamma$: age ${-}.20$ \\
\cmidrule(l){3-7}
& & \multirow{5}{*}{\checkmark} & \multirow{5}{*}{$\times$}
  & 42   & 226.43 & $\beta$: IQ ${+}.15$;\ $w$: age ${-}.22$, IQ ${+}.12$;\ $\delta$: OCI ${+}.12$ \\
& & & & 123  & 221.83 & $\beta$: IQ ${+}.15$;\ $w$: age ${-}.22$, IQ ${+}.12$;\ $\gamma$: OCI ${+}.12$ \\
& & & & 456  & 215.18 & $w$: age ${-}.17$, IQ ${+}.18$;\ bias: age ${+}.14$, OCI ${-}.13$;\ $\gamma$: OCI ${+}.14$ \\
& & & & 789  & 221.83 & $\beta$: IQ ${+}.15$;\ $w$: age ${-}.22$, IQ ${+}.12$;\ $\gamma$: OCI ${+}.12$ \\
& & & & 1234 & 226.43 & $\beta$: IQ ${+}.15$;\ $w$: age ${-}.22$, IQ ${+}.12$;\ $\gamma$: OCI ${+}.12$ \\
\cmidrule(l){3-7}
& & \multirow{5}{*}{$\times$} & \multirow{5}{*}{$\times$}
  & 42   & 230.48 & $w$: age ${-}.18$, IQ ${+}.15$;\ $\gamma$: IQ ${-}.13$ \\
& & & & 123  & 226.64 & $\beta$: IQ ${+}.15$;\ $w$: age ${-}.23$, IQ ${+}.12$;\ $\gamma$: OCI ${+}.12$;\ $\eta$: age ${+}.12$ \\
& & & & 456  & 221.83 & $\beta$: IQ ${+}.15$;\ $w$: age ${-}.22$, IQ ${+}.12$;\ $\gamma$: OCI ${-}.12$ \\
& & & & 789  & 221.83 & $\beta$: IQ ${+}.15$;\ $w$: age ${-}.22$, IQ ${+}.12$;\ $\gamma$: OCI ${-}.12$ \\
& & & & 1234 & 221.83 & $\beta$: IQ ${+}.15$;\ $w$: age ${-}.22$, IQ ${+}.12$;\ $\gamma$: OCI ${+}.12$ \\
\midrule

%% ── Learning / TD Learning ────────────────────────────────────
\multirow{20}{*}{Learning} & \multirow{20}{*}{TD Learning}
  & \multirow{5}{*}{\checkmark} & \multirow{5}{*}{\checkmark}
  & 42   & 76.06 & $\beta$: BIS ${-}.16$ \\
& & & & 123  & 70.27 & n.s. \\
& & & & 456  & 75.28 & n.s. \\
& & & & 789  & 74.84 & n.s. \\
& & & & 1234 & 75.37 & n.s. \\
\cmidrule(l){3-7}
& & \multirow{5}{*}{$\times$} & \multirow{5}{*}{\checkmark}
  & 42   & 75.37 & n.s. \\
& & & & 123  & 76.06 & $\beta$: BIS ${-}.16$ \\
& & & & 456  & 76.06 & $\beta$: BIS ${-}.16$ \\
& & & & 789  & 75.37 & n.s. \\
& & & & 1234 & 75.46 & n.s. \\
\cmidrule(l){3-7}
& & \multirow{5}{*}{\checkmark} & \multirow{5}{*}{$\times$}
  & 42   & 75.46 & n.s. \\
& & & & 123  & 75.20 & n.s. \\
& & & & 456  & 75.37 & n.s. \\
& & & & 789  & 75.46 & n.s. \\
& & & & 1234 & 70.27 & n.s. \\
\cmidrule(l){3-7}
& & \multirow{5}{*}{$\times$} & \multirow{5}{*}{$\times$}
  & 42   & 70.27 & n.s. \\
& & & & 123  & 23.21 & n.s. \\
& & & & 456  & 29.79 & n.s. \\
& & & & 789  & 75.46 & n.s. \\
& & & & 1234 & 56.48 & $\gamma$: BIS ${-}.20$ \\

\end{longtable}
\end{footnotesize}

\newpage
\section*{NeurIPS Paper Checklist}

\begin{enumerate}

\item {\bf Claims}
    \item[] Question: Do the main claims made in the abstract and introduction accurately reflect the paper's contributions and scope?
    \item[] Answer: \answerYes{} %\answerTODO{} % Replace by \answerYes{}, \answerNo{}, or \answerNA{}.
    \item[] Justification: The abstract and introduction accurately describe DisCo's components and evaluation across three tasks; claims about Validator importance are supported by ablation results in Section 4.4.

\item {\bf Limitations}
    \item[] Question: Does the paper discuss the limitations of the work performed by the authors?
    \item[] Answer: \answerYes{} %\answerTODO{} % Replace by \answerYes{}, \answerNo{}, or \answerNA{}.
    \item[] Justification: Limitations are discussed in the Conclusion, including sensitivity to the provided baseline model, reliance on task-specific validation constraints, and the need for standardized evaluation protocols for cognitive model discovery.

\item {\bf Theory assumptions and proofs}
    \item[] Question: For each theoretical result, does the paper provide the full set of assumptions and a complete (and correct) proof?
    \item[] Answer: \answerNA{} % Replace by \answerYes{}, \answerNo{}, or \answerNA{}.
    \item[] Justification: This paper proposes an empirical framework and does not include theoretical proofs or formal theorems.

    \item {\bf Experimental result reproducibility}
    \item[] Question: Does the paper fully disclose all the information needed to reproduce the main experimental results of the paper to the extent that it affects the main claims and/or conclusions of the paper (regardless of whether the code and data are provided or not)?
    \item[] Answer: \answerYes{} %\answerTODO{} % Replace by \answerYes{}, \answerNo{}, or \answerNA{}.
    \item[] Justification: Experimental details including data splits, optimization procedure, hyperparameters, prompts, discovered model programs, and hardware are described in Section 4.1 and the Appendix.

\item {\bf Open access to data and code}
    \item[] Question: Does the paper provide open access to the data and code, with sufficient instructions to faithfully reproduce the main experimental results, as described in supplemental material?
    \item[] Answer: \answerYes{} %\answerTODO{} % Replace by \answerYes{}, \answerNo{}, or \answerNA{}.
    \item[] Justification: All datasets used are publicly available from the original publications. An anonymized supplementary code package is provided for reproduction, and the code will be publicly released on GitHub upon acceptance.

\item {\bf Experimental setting/details}
    \item[] Question: Does the paper specify all the training and test details (e.g., data splits, hyperparameters, how they were chosen, type of optimizer) necessary to understand the results?
    \item[] Answer: \answerYes{} %\answerTODO{} % Replace by \answerYes{}, \answerNo{}, or \answerNA{}.
    \item[] Justification: Training/validation/test splits, optimizer (L-BFGS-B), temperature settings, number of generated candidates, search configuration, Cleaner thresholds, external validation criteria, and hardware specifications are described in Section 4.1 and the Appendix.

\item {\bf Experiment statistical significance}
    \item[] Question: Does the paper report error bars suitably and correctly defined or other appropriate information about the statistical significance of the experiments?
    \item[] Answer: \answerYes{} % Replace by \answerYes{}, \answerNo{}, or \answerNA{}.
    \item[] Justification: Results are reported across 5 independent random seeds with mean and standard deviation. BIC ranges are also provided in Table 1.

\item {\bf Experiments compute resources}
    \item[] Question: For each experiment, does the paper provide sufficient information on the computer resources (type of compute workers, memory, time of execution) needed to reproduce the experiments?
    \item[] Answer: \answerYes{} %\answerTODO{} % Replace by \answerYes{}, \answerNo{}, or \answerNA{}.
    \item[] Justification: Compute resources are specified in Section 4.1 and Appendix E: experiments used Qwen2.5-32B served with vLLM on two NVIDIA RTX A6000 GPUs (48GB VRAM each), with an AMD Ryzen Threadripper PRO 5965WX CPU (24 cores / 48 threads) and 512 GB RAM. Each seed required approximately 1--2 hours of wall-clock time, depending on the task and generated candidate models.
    \item[] Guidelines:F
    \begin{itemize}
        \item The answer \answerNA{} means that the paper does not include experiments.
        \item The paper should indicate the type of compute workers CPU or GPU, internal cluster, or cloud provider, including relevant memory and storage.
        \item The paper should provide the amount of compute required for each of the individual experimental runs as well as estimate the total compute. 
        \item The paper should disclose whether the full research project required more compute than the experiments reported in the paper (e.g., preliminary or failed experiments that didn't make it into the paper). 
    \end{itemize}
    
\item {\bf Code of ethics}
    \item[] Question: Does the research conducted in the paper conform, in every respect, with the NeurIPS Code of Ethics \url{https://neurips.cc/public/EthicsGuidelines}?
    \item[] Answer: \answerYes{} % \answerTODO{} % Replace by \answerYes{}, \answerNo{}, or \answerNA{}.
    \item[] Justification: This research analyzes existing publicly available behavioral datasets and does not involve new data collection or harmful applications.

\item {\bf Broader impacts}
    \item[] Question: Does the paper discuss both potential positive societal impacts and negative societal impacts of the work performed?
    \item[] Answer: \answerYes{} % Replace by \answerYes{}, \answerNo{}, or \answerNA{}.
    \item[] Justification: The paper discusses potential positive impacts, including reduced manual effort and theoretical bias in cognitive model construction, as well as potential negative impacts from invalid or misleading discovered models and the need for domain-grounded validation.
    
\item {\bf Safeguards}
    \item[] Question: Does the paper describe safeguards that have been put in place for responsible release of data or models that have a high risk for misuse (e.g., pre-trained language models, image generators, or scraped datasets)?
    \item[] Answer: \answerNA{} % Replace by \answerYes{}, \answerNo{}, or \answerNA{}.
    \item[] Justification: The paper does not release pretrained models or scraped datasets; all datasets are previously published and publicly available.

\item {\bf Licenses for existing assets}
    \item[] Question: Are the creators or original owners of assets (e.g., code, data, models), used in the paper, properly credited and are the license and terms of use explicitly mentioned and properly respected?
    \item[] Answer: \answerYes{} %\answerTODO{} % Replace by \answerYes{}, \answerNo{}, or \answerNA{}.
    \item[] Justification: We provide source and license or usage information for the external datasets, models, and software used in this work in Appendix~\ref{app:config}. For assets without an explicit license, we indicate that no explicit license was found and use them only for academic research.

\item {\bf New assets}
    \item[] Question: Are new assets introduced in the paper well documented and is the documentation provided alongside the assets?
    \item[] Answer: \answerNA{} %\answerTODO{} % Replace by \answerYes{}, \answerNo{}, or \answerNA{}.
    \item[] Justification: No new datasets or pretrained models are released as part of this paper.

\item {\bf Crowdsourcing and research with human subjects}
    \item[] Question: For crowdsourcing experiments and research with human subjects, does the paper include the full text of instructions given to participants and screenshots, if applicable, as well as details about compensation (if any)? 
    \item[] Answer: \answerNA{} %\answerTODO{} % Replace by \answerYes{}, \answerNo{}, or \answerNA{}.
    \item[] Justification: This paper uses existing publicly available datasets; no new human subject data was collected.

\item {\bf Institutional review board (IRB) approvals or equivalent for research with human subjects}
    \item[] Question: Does the paper describe potential risks incurred by study participants, whether such risks were disclosed to the subjects, and whether Institutional Review Board (IRB) approvals (or an equivalent approval/review based on the requirements of your country or institution) were obtained?
    \item[] Answer: \answerNA{} %\answerTODO{} % Replace by \answerYes{}, \answerNo{}, or \answerNA{}.
    \item[] Justification: No new human subjects research was conducted; all data used are from previously published studies.

\item {\bf Declaration of LLM usage}
    \item[] Question: Does the paper describe the usage of LLMs if it is an important, original, or non-standard component of the core methods in this research? Note that if the LLM is used only for writing, editing, or formatting purposes and does \emph{not} impact the core methodology, scientific rigor, or originality of the research, declaration is not required.
    %this research? 
    \item[] Answer: \answerYes{} %\answerTODO{} % Replace by \answerYes{}, \answerNo{}, or \answerNA{}.
    \item[] Justification: LLMs are a core methodological component of DisCo, used as mutation operators in the evolutionary model discovery process. Details are provided in Section 3.3.

\end{enumerate}

\end{document}